\pdfoutput=1
\pdftrailerid{}
\documentclass{fairmeta}
\usepackage{amsmath,amssymb}
\usepackage{longtable,array}
\usepackage{adjustbox}
\hypersetup{bookmarksnumbered,pdfdisplaydoctitle,pdflang={en-US},pdftitle={Hard or Mislabeled? Failure Screens Enrich Vision Benchmarks for Label Errors},pdfauthor={Haolin Guo},pdfcreator={LaTeX},pdfproducer={LaTeX}}

\crefname{figure}{Figure}{Figures}\crefname{table}{Table}{Tables}\crefname{section}{Section}{Sections}\crefname{subsection}{Section}{Sections}\crefname{appendix}{Appendix}{Appendices}\crefname{subappendix}{Appendix}{Appendices}
\title{Hard or Mislabeled? Failure Screens Enrich Vision Benchmarks for Label Errors}
\author{Haolin Guo}
\affiliation{Meta}
\abstract{Hard vision benchmarks often keep images that strong classifiers get wrong. Since a classifier is also marked wrong when the label is wrong, such failure screens may collect label errors. We measure this on ImageNet's twice-relabeled validation set with 103 classifiers. Among the 1{,}892 images that all 66 supervised classifiers get wrong, 61\% have a wrong label, versus 11.1\% overall. The fewer images a screen lets through, the more of them are mislabeled, along approximately one curve. Fit on one model family, this curve predicts the screens of two held-out model families with a mean error of 2.4 percentage points, and replays of ImageNet-D's and ImageNet-Hard's screens within 0.1 and 6.3 points. On new pools, ImageNet-V2 and CIFAR-10, enrichment at the strictest screens is stronger than the curve predicts: on ImageNet-V2, by human relabeling, 67\% of the images that all 66 classifiers get wrong are mislabeled, against 46\% predicted. Screens that let models choose among only some of the classes, as ImageNet-A's does (200 of 1{,}000), fall below the curve. By ReImageNet, 32\% of ImageNet-Hard's ImageNet images have no correct label. ScreenAudit, which comprises an audit tool, a script that rebuilds corrected answer sets for the 5{,}464 decided ImageNet-Hard images (of 5{,}646) from ReImageNet's gated release, and per-image audits of 1{,}198 benchmark images, will be released later.}
\date{November 2026}
\correspondence{Haolin Guo at \email{haolin@meta.com}}
\begin{document}
\maketitle
\begin{figure}[t]
\centering
\includegraphics[width=\linewidth]{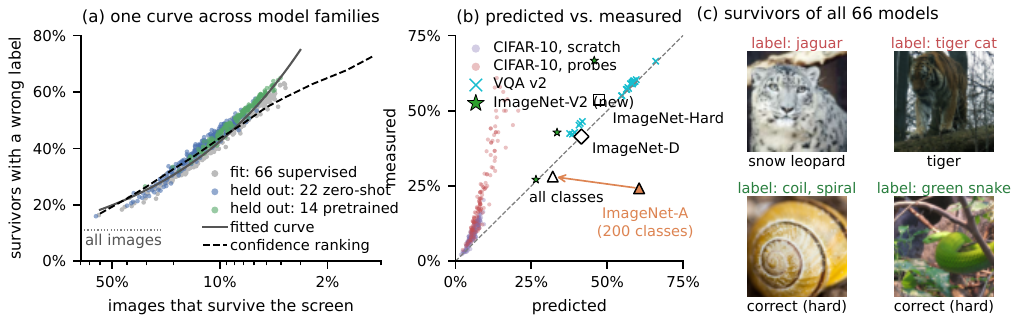}
\caption{\textbf{Failure screens collect label errors along one curve.} (a) Screens on ImageNet's validation images, scored with ReImageNet. Gray screens were used to fit the solid curve; blue and green ones come from two held-out model families. Dashed: ranking by the classifiers' mean confidence in the label (follow-up analysis). (b) The curve's prediction against the measured error rate for replays of three published screens, for CIFAR-10, for VQA v2 (there, the share of disputed answers, which marks annotator disagreement) and for three ImageNet-V2 screens (stars; human relabeling). The arrow shows ImageNet-A's screen replayed with all 1000 classes instead of its 200 (follow-up analysis). (c) Images that all 66 supervised classifiers get wrong: two wrong labels (red; relabelings' finding below) and two correct labels (green), drawn at random from the 204 correctly labeled single-object survivors that no model names correctly (follow-up analysis).}
\label{fig:teaser}
\end{figure}

\section{Introduction}
\label{sec:intro}

When a benchmark saturates, a common remedy is to keep the examples that current models still get wrong. ImageNet-A kept natural images that ResNet-50 misclassifies~\citep{hendrycks2021nae}, and ImageNet-Hard kept images that CLIP cannot classify in any of 324 zoomed crops~\citep{taesiri2023zoom}. Similar screens produced, e.g., ImageNet-D~\citep{zhang2024imagenetd}, NaturalBench~\citep{li2024naturalbench} and benchmarks that all frontier models fail~\citep{roberts2025zerobench}. In each case, the working assumption is that persistent failure marks a hard image (e.g., an unusual view or an occluded object). However, a model is also marked wrong, again and again, when the label used to score it is wrong. A screen cannot tell these two causes apart, and we expect that the stricter it is, the more of its survivors may carry a wrong label.

In this paper, we measure this effect directly. ImageNet's validation set has two independent human relabelings~\citep{beyer2020done,volkov2026reimagenet}, which lets us run any failure screen on its 50{,}000 images and count how many survivors have a wrong label. We use 103 public classifiers as filters. The share of wrong labels rises from 11.1\% of all images to 61\% of the 1{,}892 images that all 66 supervised classifiers in our fit get wrong (of 1{,}833 decided; \cref{fig:teaser}a). We call this effect \emph{error enrichment}: the fewer images survive a screen, the larger the share of wrong labels among them. Each time the share of survivors halves~\citep{guo2026trap}, the error rate among them rises by a factor of about 1.4 (1.34 when models are added, 1.49 when views are added). At a fixed share of survivors, a follow-up comparison finds that a strict screen selects wrong labels about as well as a ranking of images by the classifiers' average confidence in the label (61\% vs.\ 58\%). In this sense, these screens behave much like label-error detectors whose output is kept instead of discarded~\citep{guo2026trap}.

Because the curve depends mainly on how many images survive, it may also describe screens that were not part of its fit. We fit it on one model family and evaluate it on two held-out model families, replays of three published screens, a second image set and a new pool (\cref{fig:teaser}b). Its mean error is 2.4 percentage points for either held-out family, 0.1 for a replay of ImageNet-D's screen and 6.3 for a replay of ImageNet-Hard's zoom screen. On VQA v2, an official answer is \emph{disputed} when fewer than half of its ten annotators gave it, a sign of annotator disagreement. The curve predicts the share of disputed answers among questions that vision--language models fail within 1.6 points, and one published VQA screen within 1.3. On another, AdVQA, we expected more disputed answers than in VQA v2 and measured fewer. On new pools, enrichment at the strictest screens is stronger than predicted. On ImageNet-V2~\citep{recht2019imagenetv2}, by an independent human relabeling, 67\% of the images that all 66 classifiers get wrong are mislabeled, against 46\% predicted. On CIFAR-10, whose classifiers miss very few correctly labeled images, the mean error is 10.1 points. Screens that let models choose among only some of the classes, as ImageNet-A's does (200 of 1{,}000), fall below the curve: they let fewer images through without raising their error rate (24.1\% measured, 60.4\% predicted; \cref{sec:oos}).

Our audits show how this plays out in published benchmarks. For example, by ReImageNet, 32\% of ImageNet-Hard's ImageNet images have no correct label among the benchmark's accepted answers. One annotator's blind labels put the share of wrong labels at 41\% among its other sources (57\% after correcting for the annotator's misses). For ImageNet-A, they give 10\% (95\% CI 5--16\%), no more than ImageNet's 11.1\%, so this released benchmark shows no measurable enrichment, and our two vision--language judges find only 4.0\% and miss most of these errors. In a follow-up analysis of ImageNet-Hard's ImageNet half, we also scored models against the classes that the relabeling found in each image. This changes the best model's accuracy by 1.0 point (95\% CI $-$0.2 to 2.2) and moves top-ten models by up to 2 places (\cref{sec:audit,sec:consequences}).

These results ask for three changes, which can be made before the next benchmark is released. First, benchmark builders should replay their screen on a relabeled image set and report the error rate of its survivors. Second, model developers should report accuracy on corrected labels next to the original number. Finally, maintainers of failure-screened benchmarks should publish corrected or multi-label answer sets. Our contributions are:
\begin{itemize}
\item a controlled measurement of error enrichment in vision under 482 screens and two human relabelings, including crop and zoom effects and the role of restricted answer sets;
\item a curve fit on one model family and evaluated on held-out model families, replays of three published screens, a second image set, a new pool and VQA benchmarks, with follow-up analyses of where it misses;
\item a relabeling audit of ImageNet-Hard's ImageNet half, human and judge audits of its other sources and of ImageNet-A, and their consequences for accuracy and rankings;
\item ScreenAudit: an audit tool checked on the published screens, a script that rebuilds corrected answer sets for the 5{,}464 decided ImageNet-Hard images (of 5{,}646) from the gated ReImageNet release, per-image audits of 1{,}198 sampled benchmark images, and per-question flags of annotator disagreement for all 214{,}354 VQA v2 validation questions.
\end{itemize}

\section{Related work}
\label{sec:related}

\paragraph{Benchmarks built from model failures.}
Keeping the examples that models get wrong is a standard way to build hard test sets. ImageNet-A kept web images that ResNet-50 classifiers misclassified under ten crops, after which human annotators selected clear single-object images~\citep{hendrycks2021nae}. ImageNet-Hard kept images that CLIP ViT-L/14 misclassifies under all 324 zoomed crops, drawn from seven ImageNet-style datasets~\citep{taesiri2023zoom}. ImageNet-D kept diffusion-generated images that four models fail~\citep{zhang2024imagenetd}. Similar screens produced NaturalBench~\citep{li2024naturalbench} and MMVP~\citep{tong2024eyes}, as well as adversarial VQA sets~\citep{sheng2021advqa,li2021avqa} and benchmarks that all current models must fail~\citep{roberts2025zerobench,phan2025hle}. In language, adversarial filtering~\citep{lebras2020aflite} tends to keep examples on which annotators agree less often~\citep{phang2021adversarial}. Several of these efforts saw label problems among their survivors, which we quantify for ImageNet-Hard. For example, its authors removed 2{,}280 images with debatable labels and still estimated 14.7\% residual label noise~\citep{taesiri2023zoom}. About 40\% of NaturalBench's candidate pairs were discarded because one caption could describe both images~\citep{li2024naturalbench}, and an inspection of ImageNet-O found an in-distribution object in 20\% of its images~\citep{bitterwolf2023ninco}.

\paragraph{Label errors in vision test sets.}
ImageNet's validation labels have been reassessed several times~\citep{beyer2020done,shankar2020evaluating,tsipras2020imagenet,volkov2026reimagenet}. Test sets collected anew with ImageNet's protocol lower all models' accuracy~\citep{recht2019imagenetv2,taori2020shifts}, and roughly half of the remaining mistakes of top models are not genuine mistakes~\citep{vasudevan2022dough}. In particular, images that all models get wrong are disproportionately mislabeled or impossible~\citep{meding2022trivial}. Training dynamics separate hard from mislabeled examples~\citep{swayamdipta2020cartography}, and factor annotations attribute many model mistakes to factors (e.g., pose or object size)~\citep{idrissi2023imagenetx}. Confident learning and training-margin methods find errors by flagging images whose label a model rejects~\citep{northcutt2021confident,pleiss2020aum}, and an audit of ten test sets showed that such errors can change model rankings~\citep{northcutt2021pervasive}. ReImageNet bounded the label errors of derivative test sets using agreement between two strong models, with a lower bound of 5.8\% for ImageNet-A~\citep{volkov2026reimagenet}. Concurrent work also studies label audits~\citep{penquitt2026detection,singh2026uncertainty,lin2026atlas}. Table~S1 in the appendix summarizes what these works measured.

\paragraph{Relabeling with vision--language models.}
Today, multimodal models assist the relabeling of ImageNet~\citep{volkov2026reimagenet,kisel2026mllm} and are used to flag label errors~\citep{zhang2025lemon}. In our study, we use two such models as judges. We first measure their precision and recall against two human relabelings, on images selected by the screens we audit. One annotator's blind labels for ImageNet-A and ImageNet-Hard's other sources extend this check (\cref{sec:audit}).

A companion manuscript in preparation applies the same screen-composition analysis to the answer keys of text benchmarks~\citep{guo2026trap}.

\section{Measuring what a failure screen lets through}
\label{sec:setup}

\paragraph{Screens.}
A failure screen lets an image through when a set of filter models fails to predict its label, and we call such images \emph{survivors}. We vary screens along the axes that benchmark builders have used. First, we vary how many filters must fail, from one model to all of them, or at least $m$ of them. Second, we vary the number of attempts per filter. An attempt is either the model's standard center crop or a random resized crop with a horizontal flip, and all filters see identical random crops. Third, we use a zoom grid that reproduces the 324 crops (36 scales $\times$ a $3\times3$ grid of positions) with which ImageNet-Hard decided that an image cannot be classified~\citep{taesiri2023zoom}. As in ImageNet-A's and ImageNet-D's screens, a filter fails when its top-1 prediction differs from the dataset's original label (ImageNet-Hard judged correctness against a label set, see \cref{sec:oos}).

\paragraph{Quantities.}
For each screen, we report two numbers in percent: the share of images that survive, $f$, and the share of survivors whose label is wrong, $\theta$. We write $\pi$ for the share of wrong labels in the whole image set. Following \citet{guo2026trap}, the three are linked by $\theta=\pi c/f$, where $c$ is the share of all wrongly labeled images that survive. In other words, a screen raises the error rate whenever correctly labeled images are removed faster than wrongly labeled ones.

Following \citet{guo2026trap}, we summarize this difference by the \emph{survival ratio}: how many times more often a wrongly labeled image survives than a correctly labeled one when a filter must fail eight views in a row (median over filters).

\paragraph{Ground truth.}
As our main image set, we use ImageNet's 50{,}000 validation images~\citep{russakovsky2015imagenet}, which people have relabeled twice. ReImageNet~\citep{volkov2026reimagenet} re-annotated all images from scratch with a bounding box for each noticeable object, and we treat it as the primary truth. Under ReImageNet, an original label is wrong when its class (or an equivalent duplicate class) is not among the boxed objects, or when no ImageNet class is present at all. We set aside the 2\% of images with an unsure annotation. In addition, ReaL~\citep{beyer2020done} asked raters to accept or reject classes proposed by a set of models, and we use it as a second truth with the analogous rule. In our data, the two relabelings agree on whether the label is right for 86\% of images, so we report our results under both and note where a conclusion depends on the relabeling. Under ReImageNet, 11.1\% of the original labels are wrong (15.7\% under ReaL). A further 30\% of images show several labelled objects. We count these as correctly labeled, although a single-label benchmark may treat them as ambiguous. As a second image set, we use the CIFAR-10 test set with CIFAR-10H, which collects about 50 human votes per image~\citep{peterson2019human}. There, a label is wrong when the most common human vote is another class, which is the case for 0.8\% of images.

\paragraph{Filters.}
In total, we evaluate 103 publicly released classifiers in three families. The first has 67 convolutional networks and vision transformers trained with ImageNet labels, of which 66 enter the fit (one was added after the fit). The second has 22 image--text models used zero-shot~\citep{radford2021clip}, and the third has 14 models pretrained on up to billions of images and fine-tuned or probed with an ImageNet head. In all cases, our preprocessing follows each model's published configuration (e.g., crop size and interpolation). For the 31 held-out models with a published accuracy, our measured top-1 accuracy lies within half a point of that value, which suggests that our preprocessing matches the original. For CIFAR-10, 11 networks were trained from scratch and 8 linear probes were fitted on fixed ImageNet backbones. Weights from some sources were excluded under the study's eligibility policy, and the roster in the appendix lists the origin of each model.

\paragraph{Evaluation.}
Throughout, we fit the curve on ImageNet screens built from the first family only and evaluate it, unchanged, on screens built from the held-out families, on replays of published screens, on CIFAR-10, on ImageNet-V2 and on VQA. We set the screens, predictors and baselines before reading any held-out output, and held-out outputs were not used to fit the curve. The appendix lists every screen, predictor and baseline (Appendix~B). Code and data (ScreenAudit and the per-image files) will be released later. We label analyses added after these evaluations as follow-up analyses. Intervals are 95\% bootstrap intervals over images.

\begin{figure}[t]
\centering
\includegraphics[width=0.5\linewidth]{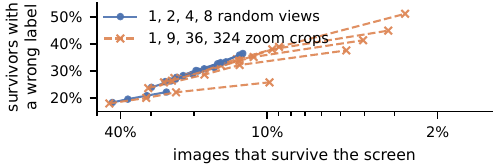}
\caption{\textbf{Zoom crops raise the error rate more slowly than random views.} Screens of five single models on ImageNet's validation images (ReImageNet); \cref{sec:controlled} gives the crop rescue counts (a follow-up analysis).}
\label{fig:insample}
\end{figure}

\section{Stricter screens let through more label errors}
\label{sec:controlled}

In \cref{fig:teaser}a, we plot each non-zoom screen built from the first family as a gray point, with the share of images that survive on the horizontal axis and the share of survivors with a wrong label on the vertical axis. We find that the error rate climbs steadily as screens become stricter, and it can exceed half of the survivors. At the mild end, a screen that keeps the images one ResNet-50 misclassifies lets through a quarter of the images, and 28\% of these have a wrong label, compared with 11.1\% of all images. We find that when all 66 classifiers must fail, 1{,}892 images survive (3.8\% of the set), and 61\% of the 1{,}833 with a decided annotation have a wrong label (69\% under ReaL). In a follow-up analysis, counting the survivors with an unsure annotation as all right or all wrong keeps this rate between 60\% and 63\%. Of these decided survivors, a further 26\% are correctly labeled but show several labeled objects. A second follow-up count finds that for 71\%, at least one of the 66 models names a class that ReImageNet finds in the image. Only 11\% are correctly labeled single-object images for which no model names a present class, which is the kind of image a hardness screen intends to find. We also find that these survivors resist new attempts. On eight fresh random views, these classifiers predict the original label for 0.7\% of them.

These data suggest that how a screen is made stricter matters less than how many images it lets through. We find that requiring more models to fail, or more views per model, moves screens along approximately one curve (Fig.~S1). A single power law in the share of survivors describes all 462 such screens with a mean absolute error of 1.4 percentage points. Still, curves fitted separately for each way of tightening a screen differ moderately. Each time the share of survivors halves, the error rate rises by a factor of 1.34 if models are added and 1.49 if views are added to one model. Again, we see these patterns under both relabelings, with a higher overall error rate and flatter growth under ReaL.

The zoom grid is the exception (\cref{fig:insample}), as its 324 crops can let both kinds of image escape. We counted these escapes for ResNet-50 in a follow-up analysis. Among the correctly labeled images that it misclassifies at the center, some crop predicts the original label for all but 10\%, regardless of object size (87\% recovered for objects under a tenth of the frame). Among the wrongly labeled ones, some crop predicts the original label for all but 22\%, presumably when a crop frames a similar object.

Wrongly labeled images thus survive the grid 2.2 times as often as correctly labeled ones, which may explain why the error rate rises more slowly under zoom (a factor of 1.24 per halving).

Finally, we find that strict screens behave much like label-error detectors~\citep{guo2026trap}. In a follow-up comparison, we ranked images by the 66 classifiers' average probability for the label and kept the lowest-ranked ones. At the share of survivors of the all-66 screen, this ranking selects 58\% wrong labels, compared with 61\% for the screen, and the two sets share 77\% of their images (\cref{fig:teaser}a).

\section{Predicting the error rate of a screen}
\label{sec:oos}

If error enrichment depends mainly on how many images survive, the curve fitted on the first family should also describe screens that it has not seen. We predict a screen's error rate from its share of survivors and the image set's error rate. Our main predictor is the power law fitted separately for each way of tightening a screen; we also report a single power law and two more flexible curves (e.g., logit-linear). We compare them with two baselines: one assumes that a screen does not change the error rate, and the other assumes that every wrongly labeled image survives. \cref{tab:oos} gives the mean absolute error of each evaluation, and \cref{fig:teaser}b shows the replays, CIFAR-10, VQA v2 and ImageNet-V2 next to their predictions.

\begin{table*}[t]
\centering
\small
\setlength{\tabcolsep}{4.5pt}\renewcommand{\arraystretch}{0.78}
\caption{\textbf{The curve on screens outside its fit.} Errors are mean absolute differences between predicted and measured shares of survivors with a wrong label, in percentage points, for the curve (with 95\% bootstrap intervals) and for the better of two baselines. For replays, the second column gives the measured and predicted error rates (\%) of the published rule itself, while the error averages the replay's screens (four nested crop budgets for ImageNet-Hard; with and without the confidence condition for ImageNet-A). ImageNet-A's rule is replayed on the 10{,}000 images of its 200 classes. For VQA v2, the predicted quantity is the share of survivors with a disputed answer (fewer than half of ten annotators gave the official answer), which marks annotator disagreement. Errors use the primary truth of each image set (ReImageNet, CIFAR-10H, VQA annotators) and, in the last column, ReaL. For ImageNet-V2, the second column gives the strictest screen, and the truth is an independent human relabeling~\citep{anzaku2024multilabelfy}.}
\label{tab:oos}
\adjustbox{max width=\linewidth}{\begin{tabular}{lcccc}
\toprule
Screens & Measured / & Error & Baseline & Error \\\\
 & predicted & (primary truth) & error & (ReaL) \\\\
\midrule
Held-out family: 22 zero-shot image--text models & 213 screens & 2.4 [1.8, 3.0] & 29.2 & 2.3 \\
Held-out family: 14 large-pretraining models & 181 screens & 2.4 [1.5, 3.5] & 39.4 & 1.0 \\
Replay: ImageNet-Hard rule (CLIP, 324 crops) & 53.6 / 47.3 & 6.6 [5.4, 7.6] & 30.0 & 2.3 \\
Replay: ImageNet-D rule (4 models) & 41.4 / 41.5 & 0.1 [0.0, 1.3] & 30.3 & 0.2 \\
Replay: ImageNet-A rule (ResNet-50, 200 classes) & 24.1 / 60.4 & 30.5 [24.8, 36.4] & 17.5 & 20.1 \\
Second image set: CIFAR-10, all 19 filters & 200 screens & 10.1 [4.9, 14.7] & 17.3 & -- \\
\quad 11 networks trained from scratch & 169 screens & 3.4 [1.3, 6.6] & 8.3 & -- \\
\quad 8 linear probes & 136 screens & 13.6 [7.4, 19.8] & 21.6 & -- \\
New pool: ImageNet-V2, 3 screens & 66.7 / 45.7 & 10.3 [8.7, 11.7] & 17.9 & -- \\
Second task: VQA v2, 6 vision--language models & 27 screens & 1.6 [1.1, 2.3] & 35.0 & -- \\
\bottomrule
\end{tabular}}
\end{table*}

\paragraph{New filters on ImageNet.}
We find that the curve fits both held-out model families closely, although the evidence comes from new filters on ImageNet's images rather than from a new image source. In particular, for 213 screens built from 22 zero-shot image--text models, the mean error is 2.4 points (95\% CI 1.8--3.0), compared with 29.2 for the better baseline. For 181 screens built from 14 large-scale pretrained models, it is 2.4 points (95\% CI 1.5--3.5), and the errors stay small under ReaL. Although these families differ from the first in training data and objectives, their screens fall on its curve (\cref{fig:teaser}a), and the single power law errs by 2.5 and 2.2 points. We ran two follow-up checks. A fitted baseline that keeps the share of surviving wrongly labeled images constant within each type of screen errs by 6.9 and 6.3 points. Similarly, refitting the rule on a random half of the images and testing it on the other half gives errors of 1.5 and 1.2 points.

\paragraph{Published screens, replayed.}
Next, three published rules were replayed on ImageNet's validation images, where the truth is known. These replays cannot measure the label quality of the released benchmarks, whose images and later manual curation differ. We find that ImageNet-D's rule, under which four models must fail, lets through 10.8\% of the images, and 41.4\% of them have a wrong label, compared with a prediction of 41.5\%. ImageNet-Hard's rule (CLIP ViT-L/14 fails all 324 crops) lets through 3.3\% of the images, of which 53.6\% have a wrong label (62\% under ReaL), 6.3 points above the prediction. Over the four zoom budgets of this replay, the mean error is 6.6 points (95\% CI 5.4--7.6) under ReImageNet and 2.3 under ReaL. Scored, as ImageNet-Hard was, against the original label plus ReaL's classes, the replay lets through 1.2\% of the images, and 56.5\% of them have a wrong label.

ImageNet-A's rule, in contrast, is far from the curve. We replayed it on the 10{,}000 validation images of its 200 classes, where 6.5\% of labels are wrong. A ResNet-50 restricted to the 200 classes, which must fail all ten crops with low confidence, lets through 2.3\% of the images. Our predicted share of wrong labels, 60.4\%, overshoots the measured 24.1\%. The original screen also used a second ResNet-50 fine-tuned on the 200 classes.

\paragraph{Why the ImageNet-A replay departs from the curve.}
In our follow-up analyses, the restricted answer set explains the miss. With all 1000 classes allowed, the same ten-crop rule lets through 6.9\% of the images, of which 28\% have a wrong label, within 4.0 points of its prediction; the restriction makes the screen look strict without making it more selective for errors (Appendix~F).

\paragraph{A second image set.}
On CIFAR-10, the strictest screens collect more wrong labels than the ImageNet curve predicts. Here, 0.8\% of the test labels disagree with the human plurality, yet the strictest screens of linear probes let through 13 to 36 images, and about half of these are mislabeled (for example, 7 of 13). The curve therefore underestimates these screens (mean error 10.1 points, 95\% CI 4.9--14.7), although it fits the screens of networks trained from scratch (3.4 points). We find that a more flexible curve, which allows steeper growth at low error rates, comes closer (5.4 points). 

CIFAR-10 also shows that a high survival ratio does not ensure a close fit. Before this evaluation, we expected the curve to carry over to an image set whose survival ratio is at least half of ImageNet's. CIFAR-10's ratio (12.3) is twice ImageNet's (5.8), yet the curve underestimates its strict screens by 10.1 points on average (13.6 for the linear probes), so the miss lies toward more enrichment than the curve implies.

A follow-up analysis suggests why this may be the case, by separating the two kinds of image that a filter misses. Single CIFAR filters let through a similar share of the wrongly labeled images as ImageNet filters do (median 0.47 vs.\ 0.53). However, they miss far fewer correctly labeled images, and the best probe misclassifies 0.8\% of the test set. These numbers suggest that correctly labeled CIFAR images may rarely defeat the tested filters, whereas the tested ImageNet filters still miss many correctly labeled images, e.g., multi-object scenes and fine-grained classes. This survival ratio records the difference (12.3 vs.\ 5.8), although two image sets cannot separate class granularity from image source or labeling policy. Among filters with matched accuracy (92 to 94.5\%), scratch networks and probes let through similar shares of wrong labels (5.0\% to 6.5\%), so the gap between the two groups appears to reflect the probes' higher accuracy.

\paragraph{A vision--language benchmark.}
To see whether error enrichment extends beyond image classification, we also evaluated the curve on VQA v2~\citep{goyal2017vqav2}. This benchmark collects ten human answers per question, and the most common one is used as the official answer. We call an official answer \emph{disputed} when fewer than half of the ten annotators gave it. Our screens score against the single official answer, as a builder who keeps one answer per question would, so a disputed answer plays the role of a wrong label, although it marks annotator disagreement rather than a verified error. Among the 50{,}000 validation questions in our sample, 19\% have a disputed answer. We screened them with 6 vision--language models. When one model fails, a median of 44\% of the surviving questions have a disputed answer. When three models fail, the median is 59\%, and when all 6 fail (13.4\% of questions survive), it is 66\%. Applied unchanged with this set's share of disputed answers, the ImageNet curve predicts the 27 screens with a mean error of 1.6 percentage points (95\% CI 1.1--2.3), compared with 35.0 for the better baseline (\cref{tab:oos}). Under VQA's own rule, where a model fails only if fewer than three annotators gave its answer, the curve errs by 7.9 points on the same screens (better baseline: 28.7) and underestimates the strictest one (77\% vs.\ 67\%; a follow-up analysis). Unlike the CIFAR-10 filters, each of our models still fails roughly a third of all questions, many of which have an undisputed answer. We also applied the curve to two published VQA screens (Appendix~H). For VQA-CE~\citep{dancette2021vqace}, which keeps questions that shortcut rules fail, it predicts 41.5\% disputed answers against 40.2\% measured, an error of 1.3 points (better baseline: 21.0). For AdVQA~\citep{sheng2021advqa}, which kept questions that fooled models, we expected a higher share than in VQA v2 but measured a lower one, 10.5\% against 19.2\% (difference 95\% CI -9.3 to -8.1 points). Its builders also removed questions that annotators judged invalid (an explanation offered after the result).

\paragraph{A new pool.}
Finally, we built three screens on ImageNet-V2~\citep{recht2019imagenetv2}, a 10{,}000-image replication of ImageNet's validation set on which the curve had not been evaluated: one ResNet-50 fails, at least 50 of the 66 fitted classifiers fail, or all 66 fail. The predictions use the pool's error rate, which we measured with an independent human relabeling~\citep{anzaku2024multilabelfy}. Scored with that relabeling, the curve's mean error is 10.3 points (95\% CI 8.7--11.7), against 17.9 for the better baseline. It predicts the mildest screen closely (27\% measured, 27\% predicted) but underestimates the stricter ones: 43\% against 33\%, and 67\% against 46\% for the 726 images that all 66 classifiers get wrong, as on CIFAR-10. With our two judges as truth, on a sample of 2{,}437 images, the mean error is 11.2 points (95\% CI 6.1--31.8) against 20.7 for the better baseline.

Taken together, on ImageNet's validation images the curve describes the held-out families' screens and the ImageNet-D and ImageNet-Hard replays within a few points, and it predicts the share of disputed answers on VQA v2 and VQA-CE. It overestimates a screen that restricts the answer set and underestimates the strictest screens on CIFAR-10 and ImageNet-V2, so on a new pool it should be recalibrated on a labeled sample of survivors (\cref{sec:recipe}).

\section{Auditing ImageNet-A and ImageNet-Hard}
\label{sec:audit}

\paragraph{Direct measurement.}
Half of ImageNet-Hard's 10{,}980 images come from ImageNet. For these images, we can measure label errors without any judge. To this end, we matched 5{,}646 of them to their validation originals by comparing $32\times32$ thumbnails with the validation images of their class, keeping only matches with identical image dimensions. Here, a thumbnail correlation above 0.95 marks an unchanged copy (4{,}551 images), and one between 0.6 and 0.95 a corrupted variant (1{,}095 images). For 99\% of the matches, the best candidate exceeds the second-best by more than 0.05. The relabelings describe the objects in the original scene, and they are applied to both kinds of copy.

In our audit, we say that an image has \emph{no correct label} when none of the classes that ImageNet-Hard accepts for it, after adding equivalent class names, is among the objects that ReImageNet's annotators boxed, or when they found no ImageNet class at all. By this definition, 32\% (95\% CI 31.2--33.7) of the 5{,}464 matched images with a decided annotation have no correct label, 33.5\% of the unchanged copies and 28.1\% of the corrupted variants.

ReaL tells a consistent story, although it can miss errors that no model proposed. Its raters had accepted none of the proposed classes for 39\% of the matched images, compared with 6\% of ImageNet's validation set as a whole. ImageNet-Hard accepts ReaL's classes as answers in addition to the original label, but for 93\% of the matched images it accepts only the original label. The benchmark's authors estimated 14.7\% residual label noise by an undescribed manual inspection~\citep{taesiri2023zoom}, so the two numbers are not directly comparable.

\begin{table}[t]
\centering
\small
\setlength{\tabcolsep}{3pt}\renewcommand{\arraystretch}{0.82}
\caption{\textbf{Judge validation} on ImageNet validation images against ReImageNet. An image counts as called mislabeled when both judges say so; the last two columns score ReaL itself against ReImageNet on the same images. Wrong: share of images with a wrong label; Called: share called mislabeled (both in \%). Images with a decided ReImageNet annotation only.}
\label{tab:judge}
\adjustbox{max width=\linewidth}{\begin{tabular}{lrrrcccc}
\toprule
 & & & & \multicolumn{2}{c}{Judges} & \multicolumn{2}{c}{ReaL} \\
Images & $n$ & Wrong & Called & Prec. & Rec. & Prec. & Rec. \\
\midrule
Random & 292 & 11.0 & 7.5 & 0.68 & 0.47 & 0.36 & 0.56 \\
All 8 filters fail & 290 & 50.7 & 40.3 & 0.83 & 0.66 & 0.69 & 0.82 \\
ImageNet-Hard rule & 196 & 49.0 & 39.3 & 0.90 & 0.72 & 0.66 & 0.81 \\
Screen-selected & 486 & 50.0 & 39.9 & 0.86 & 0.68 & 0.68 & 0.81 \\
\bottomrule
\end{tabular}}
\end{table}

\paragraph{Judge audit.}
No independent relabeling exists for ImageNet-A or for ImageNet-Hard's other sources (e.g., sketches). For these images, we therefore rely on two vision--language judges (Claude Opus 5.5 and GPT-6.1 Sol; Appendix~D), whose calls may be wrong in either direction. We count an image as mislabeled when both judges say so. On ImageNet validation images selected by the screens we study, this joint call has a precision of 0.86 and a recall of 0.68 when scored with ReImageNet (\cref{tab:judge}). These values exceed the thresholds that we required of the judges (precision at least 0.80, recall at least 0.50). Hiding the model-suggested options changes these values little (Appendix~F).

In our checks, we observe that the judges miss many wrong labels and that their errors depend on the kind of image. On random images, their precision drops to 0.68, and they disagree with each other on 14\% of screen-selected images. In addition, we checked them on the 301 ImageNet-sourced images of our 598-image ImageNet-Hard sample that have a decided direct truth. There, they call 17\% mislabeled while the true rate is 32\% (precision 0.76, recall 0.40).

By comparison, ImageNet-Hard's other sources are called mislabeled more often than its ImageNet half. Both judges call 30.5\% (95\% CI 25.1--36.0) of the 275 sampled images from its sketch, ObjectNet~\citep{barbu2019objectnet} and three smaller sources (e.g., renditions) mislabeled, compared with 17\% of its ImageNet-sourced images. Corrected with the judges' error rates on four ImageNet image sets, this gives 32\% to 72\%. Because these error rates come from ImageNet images, 200 of these images were also labeled blind by one annotator\footnote{The annotator, an adult who participated voluntarily, saw only public benchmark images, their labels and neutral written instructions, and no model outputs. No IRB review was carried out and no written consent was sought, because the task posed minimal risk and collected no personal data.} (Appendix~E). Of the 143 images that this single annotator could decide, 41\% have a wrong label (95\% CI 33--48\%; e.g., sketches 34\%, ObjectNet 66\%; judge-calibrated estimate 39\%). Correcting for the annotator's misses on known-truth controls (4 of 14) gives 57\% (40--90\%; a follow-up analysis). Scored on these labels, the judges' joint call has a precision of 0.84 and a recall of 0.65. However, ObjectNet's labels come from how its photos were taken, so a ``wrong'' there may partly mean an object that cannot be recognized. Overall, weighted by source size, 36\% (32--40\%) of ImageNet-Hard's images have a wrong label without this correction and 44\% (36--59\%) with it; the measured ImageNet half contributes 17 points.

Finally, both judges call 4.0\% of ImageNet-A's 600 sampled images mislabeled (95\% CI 2.5--5.5; at least one judge: 15.8\%). Corrected with the same four calibrations, this share gives 0\% to 3\%, below the lower bound of 5.8\% from ReImageNet's agreement analysis~\citep{volkov2026reimagenet}, so these calibrations do not transfer to this benchmark. These blind labels confirm this: of the 140 decided images, 10\% have a wrong label (95\% CI 5--16\%; 14\% after the same correction, 7--25\%), and the judges' joint call finds only one of these (recall 0.07). This share is no higher than ImageNet's 11.1\% (6.5\% for the same 200 classes, inside its interval), so the released ImageNet-A shows no measurable enrichment. Since 60 of these 200 images were marked unclear, counting all of them as correct or as wrong bounds the share at 7--37\%.

\section{What corrected labels change}
\label{sec:consequences}

Wrong labels lower the measured accuracy of models on a failure-screened benchmark (Table~S10 in the appendix). Strong models may be affected most, because more of their errors can fall on images without a correct label. We score ImageNet-Hard's ImageNet half in two ways, and we report both. First, lenient scoring adds ReImageNet's classes to the accepted answers and keeps the original label. Under it, the best of our 103 models goes from 62.5\% to 70.5\% (8.0 points, 95\% CI 7.2--8.7), and about half of its original errors (51\%) fall on images without a correct accepted answer. However, lenient scoring still credits wrong original labels. A follow-up strict scoring therefore accepts only the classes that ReImageNet found in the image, on the 4{,}767 images (87\%) that contain an ImageNet class. On these images, the same model goes from 64.3\% to 65.3\% (1.0 point, 95\% CI $-$0.2 to 2.2). It gains 26\% of its original errors but loses 13\% of its original successes, which had been credited to absent classes.

On our full 598-image sample, lenient scoring with the class that both judges name raises the best model's accuracy by 10.5 points (95\% CI 8.0--13.0). Since our judges miss about a third of the wrong labels that our annotator finds, this may understate the change.

We find that these corrections change rankings less than accuracy. Specifically, Kendall's $\tau$ between the original and corrected rankings (1 means identical) is 0.90 (95\% CI 0.86--0.92) on the ImageNet-Hard sample. Under strict scoring of the ImageNet half, it is 0.93, and top-ten models move by up to 2 places. On ImageNet-A, where the judges miss almost all errors that our annotator finds, judge-corrected scores change little (Appendix~F).

Corrections can also change a robustness reading. Under the whole benchmark's corrected answers, the effective robustness~\citep{taori2020shifts} of the 22 zero-shot image--text models on ImageNet-Hard moves from $-$3.7 (95\% CI $-$4.2 to $-$3.2) to $+$2.1 points (95\% CI 1.5--2.7). On the human-corrected ImageNet half, it moves from $-$12.5 to $-$8.4 and keeps its sign. Outside the ImageNet half, these corrections come from the judges, whose candidate classes were proposed in part by an image--text model (Appendix~F).

\section{A screen audit for benchmark builders}
\label{sec:recipe}

We turn our results into a tool that builders can run before release. Given the survivors of a screen on ImageNet's validation images, it reports the share of survivors with a wrong label under both relabelings, with intervals, and the curve's prediction. When the screen restricts the answer set, the tool warns that the prediction is unreliable and, given the survivors of the same screen replayed with all classes, reports that replay too. Given only the share of candidate images that survive and a rough error rate for the candidate pool, it returns the curve's prediction with a range of $\pm$4.0 points. This range is the 90th percentile of the errors on 399 held-out screens of ImageNet's validation images (394 new-family screens and five replays), so it may not hold for other pools. On the published screens, the prediction misses by 0.1 points for ImageNet-D and 6.3 for ImageNet-Hard, but by 36.3 for ImageNet-A's 200-class screen (4.0 with all classes), by 10.1 on CIFAR-10 and by 10.3 on ImageNet-V2. A new pool therefore needs recalibration on a labeled sample of survivors. The tool comes with a script that rebuilds corrected answer sets for the 5{,}464 decided ImageNet-Hard images (of 5{,}646) from the gated ReImageNet release, together with the match and copy type of each image, the judges' calls for 1{,}198 sampled images, per-question flags of annotator disagreement for all 214{,}354 VQA v2 validation questions (41{,}201 disputed) and six models' answers on a 50{,}000-question sample. We call the tool and these files ScreenAudit. Code and data (ScreenAudit and the per-image files) will be released later.

In practice, a builder can replay the screen on a fully relabeled image set, with the full answer set if the screen restricts answers, and report accuracy under the original and corrected answers. Where such a relabeling exists, the survivors can also be verified before release. In a follow-up analysis, we kept the 5{,}735 validation images that at least 50 of our 66 classifiers get wrong. Our 36 held-out models reach 22\% accuracy on them, and about half of their errors (48\%) fall on wrong labels. Both relabelings agree that 444 of these images show only their labeled class. On these verified images, the same models reach 44\%, so the set stays hard (Appendix~F). On ImageNet-V2, the same procedure gives 23\% accuracy on the screened images and 45\% on the 265 verified ones. Without a relabeling, a random sample of the survivors can be audited with two judges calibrated on the same kind of images. Beyond classification, the same audit could serve failure-screened multimodal benchmarks (e.g., NaturalBench~\citep{li2024naturalbench} and ZeroBench~\citep{roberts2025zerobench}), where answer keys play the role of labels.

\section{Limitations}
\label{sec:discussion}
Our two relabelings disagree on 14\% of ImageNet's images, so we report both. The curve was fitted on one image set. The held-out families test new models on the same images, and on CIFAR-10 and ImageNet-V2 the curve underestimates the strictest screens, so a new pool needs recalibration. The audit of non-ImageNet sources rests on two closed, versioned judges whose recall varied from 0.40 to 0.72 across ImageNet image sets. Our human labels come from a single annotator, who marked 28\% of items unclear and, on known-truth controls, was right on 86\%, missing 29\% of wrong labels but flagging no correct ones. All filters see identical random views, and the ImageNet-Hard replay uses the original label as its only correct answer. We count a present original class as correct even in multi-object scenes, so our error rates may understate problems for single-label scoring. The appendix gives details.

\section{Conclusion}
\label{sec:conclusion}
Keeping the images that models get wrong collects wrong labels, and the stricter the screen, the larger their share. On ImageNet, this error enrichment follows one curve across model families, and the curve predicts the screens of held-out model families on ImageNet's validation images within a few points. It overestimates a screen that restricts the answer set and underestimates the strictest screens on CIFAR-10 and ImageNet-V2; follow-up analyses trace the first two misses to the restricted answer set and to filters that rarely miss correctly labeled images. Since strict screens behave much like label-error detectors~\citep{guo2026trap}, we suggest that the images they flag be relabeled before the rest are read as hard.

We hope that the next failure-screened benchmark reports the error rate of its survivors next to its accuracy table.

\section*{LLM Usage Statement}
LLM-based coding agents designed, implemented and ran the experiments and analyses under the author's direction and review. Language models used as study subjects, judges or material writers are described in the methods. The author wrote the text; LLMs were used only for grammatical edits. The author is responsible for all content. Tools: Claude Opus 5.5 in an agentic coding harness (experiment, analysis and figure code, and running the experiments) and GPT-6.1 Sol via the OpenAI Codex CLI (reviews of the study design and the drafts); grammatical edits used Claude Opus 5.5.

\clearpage
\beginappendix
\setcounter{table}{0}\setcounter{figure}{0}
\renewcommand{\thetable}{S\arabic{table}}\renewcommand{\thefigure}{S\arabic{figure}}
\section{Prior work}
\label{sec:supp_prior}
\cref{tab:supp_prior} summarizes what related work measured.

\begin{table*}[!h]
\centering\small
\setlength{\tabcolsep}{4pt}
\begin{adjustbox}{max width=\linewidth}\begin{tabular}{@{}lcccccc@{}}
\toprule
 & Errors in & Grows with & Predicts & Vision screens & Audits of & Accuracy, \\
 & screened sets & strictness & new screens & and mechanisms & vision benchmarks & ranks \\
\midrule
Relabelings~\citep{beyer2020done,volkov2026reimagenet} & bound & -- & -- & -- & bound & \checkmark \\
Top-model errors~\citep{vasudevan2022dough,meding2022trivial} & \checkmark & -- & -- & -- & -- & \checkmark \\
Error detectors~\citep{northcutt2021confident,pleiss2020aum,zhang2025lemon} & -- & -- & -- & -- & -- & -- \\
Pervasive label errors~\citep{northcutt2021pervasive} & -- & -- & -- & -- & \checkmark & \checkmark \\
Adversarial filtering~\citep{lebras2020aflite,phang2021adversarial} & agreement & -- & -- & -- & -- & -- \\
Screened benchmarks~\citep{hendrycks2021nae,taesiri2023zoom} & manual & -- & -- & -- & -- & -- \\
Ours & \checkmark & \checkmark & \checkmark & \checkmark & \checkmark & \checkmark \\
\bottomrule
\end{tabular}\end{adjustbox}
\caption{\textbf{What prior work measured.} Bound: an agreement-based lower bound; agreement: lower annotator agreement among kept items; manual: the builders' own estimate after manual cleaning.}
\label{tab:supp_prior}
\end{table*}

\section{Protocol, code and deviations}
\label{sec:supp_plan}
The experiment code and analyses were written and run by Claude Opus 5.5 in an agentic coding tool under the author's direction; GPT-6.1 Sol, via the OpenAI Codex CLI, reviewed the study design and the drafts, and both models compared draft versions blind during revision. The order of work was: in-sample inference and pilots; a design review; CIFAR-10 training, with all 20 models' test outputs stored with checksums; fitting the curve on the in-sample family; held-out inference; and the held-out evaluations.\footnote{Evaluation labels used in the code: P1 zero-shot family; P2 large-pretraining family; P3 ImageNet-Hard replay; P4 ImageNet-D replay; P5 ImageNet-A replay; P6 CIFAR-10; P7 survival-ratio check; P8 the VQA v2 evaluation.} Before any held-out output was read, we set the primary truth (ReImageNet; ReaL secondary), the in-sample family, the screens of each evaluation, the four predictors and two baselines, the judge protocol with its acceptance thresholds, the audit samples and the annotation procedure. Following the design review, we compute error rates over decided images only, use truly nested zoom budgets, compare competing predictors with bootstrapped baselines, give the judges equivalent-class label sets, and add a candidate-anchoring ablation and a human-label calibration.

\paragraph{Code and analysis files.} Code and data (ScreenAudit and the per-image files) will be released later. Six code files were edited after the curve was fit, for logged reasons.

\paragraph{Deviations.} We logged every deviation before the affected outputs were read.
(i) One in-sample model (EfficientFormerV2-L) was still running when the curve was fit and was excluded from the fit.
(ii) The OpenAI CLIP checkpoints had first been loaded without their QuickGELU activation; we fixed the configuration and re-ran all affected outputs before any was analyzed.
(iii) Two of 600 ImageNet-Hard audit images could not be judged within the time limit (598 judged).
(iv) The human annotation was amended three times before any label existed: one annotator instead of two, a threshold of 190 answered items per source, and 30 known-truth items (\cref{sec:supp_humans}).
(v) The VQA v2 evaluation was designed after the image evaluations and before any VQA output; one amendment replaced models that did not meet the format rule (\cref{sec:supp_vqa}).
(vi) Llama-3.2-11B-Vision weights were loaded from an unmodified community mirror; the model was dropped by the format rule.
(vii) A re-run of ImageNet-A's ten-crop outputs used in two follow-up tables differs from the main replay by one image (223 vs.\ 224 survivors).
(viii) We expected 5{,}768 same-size matches for ImageNet-Hard's ImageNet half; the audit also requires a thumbnail correlation above 0.6, which leaves 5{,}646.
(ix) An independent audit found that an earlier strict-scoring comparison averaged the original accuracy over different images; the paper reports the corrected like-for-like comparison (\cref{sec:supp_followups}).
Operational changes (job order, image sharding, additional GPU machines, a restarted VQA run) did not alter any analysis.

\section{Screens and forecasting rules}
\label{sec:supp_rules}
\paragraph{Screen types.} With filters $\mathcal{F}$ and attempts indexed by $j$ ($j=0$ the center view, $j=1,\dots,7$ random views used for selection), a filter $m$ fails image $x$ at depth $k$ when its top-1 prediction differs from the original label on attempts $0,\dots,k-1$. We use five types: one filter at depth $k\in\{1,2,4,8\}$ (``one''); all of $n$ filters at the center view (``all\_n''); all of $n$ filters at depth $k$ (``all\_n\_depth''); at least $m$ of all filters (``at\_least\_m''); and a zoom budget of 1, 9, 36 or 324 nested crops (``zoom''). Random views 8--15 are reserved for measuring whether survivors are solved on fresh attempts. All filters see the same random views.

\paragraph{Zoom budgets.} The grid has 36 scales (10 to 1024 pixels) and a $3\times3$ grid of anchors per scale, giving 324 crops that are resized to each model's input size. The nested budgets are: 1 crop (scale 224, center anchor); 9 crops (scale 224, all anchors); 36 crops (scales 224, 235, 240 and 256, all anchors); and all 324 crops.

\paragraph{CIFAR-10 filters.} The 11 networks trained from scratch are ResNet-18 (two seeds), ResNet-34, a narrow ResNet-18, VGG-11, VGG-16, WRN-16-8, WRN-28-4, ConvMixer-256/8, a small ViT and a VGG-11 trained for 8 epochs. Training used 40 epochs (60 for the ViT), SGD with Nesterov momentum or AdamW (ViT), a one-cycle schedule, batch size 512, random crops with 4-pixel padding and flips. A twelfth network was trained but excluded from the filter set beforehand. The 8 linear probes are L-BFGS logistic regressions on fixed features of ImageNet backbones from the in-sample family. Test accuracies range from 83\% to 94\% for scratch networks and from 92\% to 99\% for probes. All 20 trained models' test outputs were stored with checksums before the curve was fit and read only for the CIFAR-10 evaluation.

\paragraph{Quantities.} For a screen $S$ on a pool with truth $T$, $f=|S\cap\mathcal{R}|/|\mathcal{R}|$ and $\theta=|S\cap\mathcal{I}|/|S\cap\mathcal{R}|$, where $\mathcal{R}$ is the set of decided images and $\mathcal{I}$ the wrongly labeled ones. The share of wrongly labeled images that survive is $c=|S\cap\mathcal{I}|/|\mathcal{I}|$, so $\theta=\pi c/f$~\citep{guo2026trap}.

\paragraph{Rules.} R1 (primary) fits $\log c = a_t + \beta_t \log f$ separately for each type of screen $t$; R1g fits one law for all types; R2 fits $\operatorname{logit}\theta$ linearly in $\log f$ per type; R3 is an isotonic (monotone) fit of $\theta$ on $\log f$ per type. The forecast for a held-out screen is $\hat\theta=\min(1,\pi\,e^{a_t}f^{\beta_t-1})$ for R1, with $\pi$ measured on the held-out pool. The baselines are $\hat\theta=\pi$ (the screen does not change the error rate) and $\hat\theta=\min(1,\pi/f)$ (every wrongly labeled image survives). Screens with fewer than 30 decided survivors (10 for CIFAR-10) are not scored. Intervals use 500 Poisson bootstrap draws over the held-out images, with the rules held fixed. \cref{tab:supp_mult} lists the fits and \cref{tab:inventory} the scored screens.

\paragraph{Survival ratio.} The survival ratio is the median over filters of how many times more often a wrongly labeled image than a correctly labeled one survives eight failed views. It is 5.8 on ImageNet, 5.1 and 7.5 for the two held-out families, and 12.3 on CIFAR-10, where the curve underestimates the strict screens; a high ratio therefore does not ensure a close fit.

\begin{table*}[t]
\centering\small
\setlength{\tabcolsep}{4pt}
\begin{adjustbox}{max width=\linewidth}\begin{tabular}{@{}lrrrrrr@{}}
\toprule
Screens built from & Total & one model, $k$ views & all $n$ models fail & $n$ models, $k$ views & $\geq m$ of $n$ fail & zoom grid \\
\midrule
Fit: 66 supervised (ImageNet) & 482 & 264 & 101 & 90 & 7 & 20 \\
22 zero-shot (ImageNet) & 213 & 88 & 61 & 60 & 4 & 0 \\
14 large-pretraining (ImageNet) & 181 & 56 & 61 & 60 & 4 & 0 \\
ImageNet-Hard replay & 4 & 0 & 0 & 0 & 0 & 4 \\
ImageNet-D replay & 1 & 0 & 1 & 0 & 0 & 0 \\
ImageNet-A replay & 2 & 2 & 0 & 0 & 0 & 0 \\
CIFAR-10, 11 scratch nets & 169 & 44 & 61 & 60 & 4 & 0 \\
CIFAR-10, 8 probes & 136 & 32 & 41 & 59 & 4 & 0 \\
\bottomrule
\end{tabular}\end{adjustbox}
\caption{\textbf{Screen inventory.} Scored screens (at least 30 decided survivors; 10 on CIFAR-10) per test and kind. Model subsets for multi-model screens are drawn at random with the seeds listed in the code. The CIFAR-10 test pools both groups (200 screens); 169 and 136 screens are scored within each group.}
\label{tab:inventory}
\end{table*}

\begin{table}[t]\centering\small\setlength{\tabcolsep}{3pt}
\caption{In-sample fits of $\log c = a + \beta \log f$ by screen type, with the implied factor $2^{1-\beta}$ by which the error rate grows when the share of survivors halves.}
\label{tab:supp_mult}
\adjustbox{max width=\linewidth}{\begin{tabular}{lrcccccc}\toprule
 & & \multicolumn{3}{c}{ReImageNet} & \multicolumn{3}{c}{ReaL}\\
Screen type & $n$ & $a$ & $\beta$ & factor & $a$ & $\beta$ & factor\\\midrule
one model, more views & 264 & 0.070 & 0.428 & 1.49 & 0.426 & 0.625 & 1.30 \\
more models fail (center view) & 101 & 0.389 & 0.581 & 1.34 & 0.681 & 0.747 & 1.19 \\
more models and views & 90 & 0.429 & 0.598 & 1.32 & 0.693 & 0.748 & 1.19 \\
at least $m$ of 66 fail & 7 & 0.187 & 0.469 & 1.45 & 0.394 & 0.596 & 1.32 \\
zoom grid & 20 & 0.377 & 0.685 & 1.24 & 0.522 & 0.731 & 1.21 \\
all non-zoom screens (single law) & 462 & 0.259 & 0.528 & 1.39 & 0.552 & 0.693 & 1.24 \\
\bottomrule\end{tabular}}\end{table}

\begin{table*}[t]\centering\small\setlength{\tabcolsep}{3pt}
\caption{Held-out errors (mean absolute error, percentage points) of all four predictors under both truths. R1: type-matched power law (the curve used in the paper); R1g: single power law; R2: logit-linear; R3: monotone; Base: better of the two baselines.}
\label{tab:supp_truths}
\adjustbox{max width=\linewidth}{\begin{tabular}{l ccccc ccccc}\toprule
 & \multicolumn{5}{c}{ReImageNet} & \multicolumn{5}{c}{ReaL}\\
Screens & R1 & R1g & R2 & R3 & Base & R1 & R1g & R2 & R3 & Base\\\midrule
Zero-shot family (22) & 2.4 & 2.5 & 2.3 & 2.4 & 29.2 & 2.3 & 2.2 & 2.3 & 2.3 & 33.2 \\
Large-pretraining family (14) & 2.4 & 2.2 & 2.3 & 2.1 & 39.4 & 1.0 & 0.9 & 0.9 & 0.9 & 40.0 \\
ImageNet-Hard rule & 6.6 & 5.1 & 6.5 & 6.0 & 30.0 & 2.3 & 6.4 & 1.3 & 1.0 & 34.7 \\
ImageNet-D rule & 0.1 & 0.4 & 0.3 & 0.3 & 30.3 & 0.2 & 0.8 & 0.0 & 0.1 & 38.7 \\
ImageNet-A rule & 30.5 & 21.8 & 33.3 & 13.1 & 17.5 & 20.1 & 13.4 & 25.0 & 7.8 & 30.8 \\
\bottomrule\end{tabular}}\end{table*}

\section{Truths, replays and judges}
\label{sec:supp_judges}
\begin{figure*}[t]\centering\includegraphics[width=\linewidth]{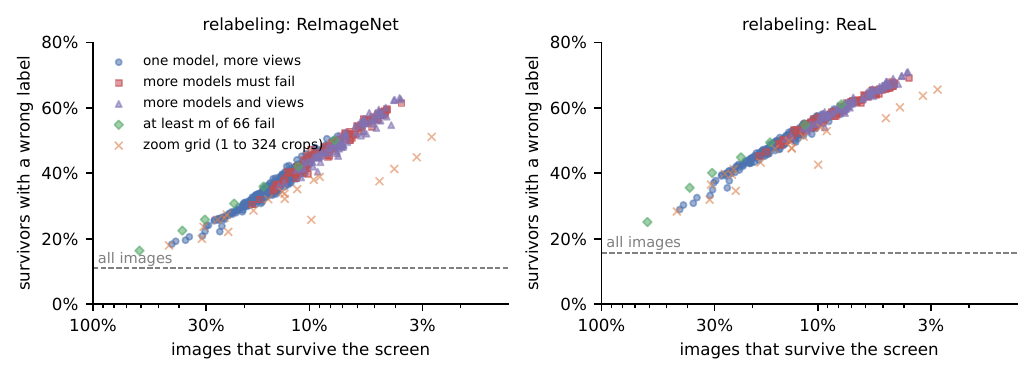}
\caption{\textbf{In-sample screens under both relabelings.} Screens built from the 66 supervised classifiers; dashed lines: share of wrong labels among all images. Screens made stricter with more models, more views or both fall approximately on one curve, while zoom screens rise more slowly.}\label{fig:supp_truths}\end{figure*}
\paragraph{Agreement of the two truths.} \cref{fig:supp_truths} shows the in-sample screens under both relabelings. On the 48{,}912 decided validation images, ReImageNet and ReaL agree on whether the label is right for 86\%: 3{,}008 images are wrongly labeled under both, 2{,}402 under ReImageNet only and 4{,}448 under ReaL only. ReaL marks more labels wrong mainly because its raters accepted none of the model-proposed classes for 6\% of images, while ReImageNet's annotators boxed any noticeable object. Counting survivors with an unsure annotation as all right or all wrong bounds the all-66 screen's error rate between 60\% and 63\% and the ImageNet-Hard replay's between 52\% and 55\%.

\paragraph{Replay fidelity.} The ImageNet-D replay uses CLIP ViT-L/14, ViT-L/14-336, RN50 and a torchvision ResNet-50, all of which must misclassify the center view. The ImageNet-Hard replay uses CLIP ViT-L/14 with the 324-crop grid and scores against the original label; ImageNet-Hard itself scored against the original label plus ReaL's classes, and with that criterion 1.2\% of the images survive, of which 56.5\% have a wrong label (50\% have no ReaL class at all). The ImageNet-A replay uses one torchvision ResNet-50 with logits restricted to the 200 classes, ten crops, and a probability of at most 0.15 for the label on all crops; the original screen also used a ResNet-50 fine-tuned on the 200 classes, capped confusion pairs and ended with manual selection, none of which the replay reproduces.

\paragraph{Audit samples.} The ImageNet-A sample has 600 images drawn at random (seed 1104) from its 7{,}500. The ImageNet-Hard sample has 600 images drawn at random (seed 1105) from its 10{,}980: 323 from ImageNet, 180 from ImageNet-Sketch, 77 from ObjectNet, 8 from ImageNet-R, 6 from ImageNet-A and 6 from ImageNet-V2. Two ObjectNet images could not be judged within the time limit, leaving 598. Of the 323 ImageNet-sourced images, 301 were matched to a validation image with a decided ReImageNet annotation; the other 22 are counted in the judges' rates but not in the direct check. Benchmark-wide rates weight the ImageNet half (5{,}890 images) by its directly measured rate and the other sources (5{,}090 images) by their judge-based range.

\paragraph{Judge protocol.} Each judge sees one image and a short list of options. Each option is an ImageNet class given by its names and a dictionary definition. The options are the benchmark's label set (with the equivalent class of each of 14 duplicated ImageNet class pairs) and up to four alternatives proposed by three classifiers from three families (the top-2 of BEiTv2-L and the top-1 of DFN5B-H/14 and of DINOv2-g with a linear head), deduplicated and shuffled per image. Images with a long side above 1024 pixels are downscaled. The instruction reads:
\begin{quote}\small
ImageNet treats its classes as separate categories, each defined below. A label is correct only if the image clearly shows an object that belongs to THAT category. It is wrong if the object belongs to a different category instead, including a more specific or a look-alike category among the options (for example, a bolete is not the generic ``mushroom'' category, a golden retriever is not a cocker spaniel, a monitor is not a ``screen, CRT screen''). If the image shows several objects, the label is correct when any clearly visible object belongs to the label's category.
\end{quote}
The judge answers \emph{correct}, \emph{wrong} (naming the best option, if listed), \emph{no\_class} or \emph{unclear}, and lists other options that are visible. An image counts as mislabeled when both judges answer wrong or no\_class, as correctly labeled when both answer correct, and as undecided otherwise; undecided counts as correctly labeled in all rates. Both judges ran with fixed settings (high reasoning effort, no tools, one call per image, a 600-second limit and up to three attempts). The judges are Claude Opus 5.5 ({\tt claude-opus-5-5}) and GPT-6.1 Sol ({\tt gpt-6.1-sol}), closed commercial models whose identifiers are fixed in the judge code. \cref{tab:supp_judges} gives the validation by stratum and by judge. The validation tables count images with a decided ReImageNet annotation only (292, 290 and 196 of 300, 300 and 200).

\paragraph{Checking the correction on known truth.} Applied to the 323 ImageNet-sourced images of the ImageNet-Hard sample, whose direct error rate is 32.4\%, the four calibrations give 32.4\% (random images), 5.8\% and 14.1\% (the two screen-selected sets) and 32.5\% (the ImageNet-Hard check, which is fitted on these images). The screen-selected calibrations thus underestimate the error rate on these images, so the lower end of the 32--72\% range for the other sources is likely an underestimate.

\begin{table}[t]\centering\small\setlength{\tabcolsep}{2.5pt}
\caption{Judge validation by stratum against ReImageNet: joint decision with 95\% bootstrap intervals, each judge alone (precision / recall), and Cohen's $\kappa$ between judges.}
\label{tab:supp_judges}
\adjustbox{max width=\linewidth}{\begin{tabular}{lccccc}\toprule
Images & Precision & Recall & Claude & GPT & $\kappa$\\\midrule
Random & 0.68 [0.48, 0.88] & 0.47 [0.28, 0.65] & 0.69 / 0.56 & 0.49 / 0.56 & 0.56 \\
All 8 filters fail & 0.83 [0.75, 0.89] & 0.66 [0.58, 0.73] & 0.79 / 0.73 & 0.79 / 0.75 & 0.64 \\
ImageNet-Hard rule & 0.90 [0.82, 0.96] & 0.72 [0.62, 0.81] & 0.87 / 0.77 & 0.80 / 0.77 & 0.69 \\
\bottomrule\end{tabular}}\end{table}

\section{Human labels}
\label{sec:supp_humans}
\paragraph{Design.} One adult volunteer annotator, arranged by the author, labeled 450 items in a private answer sheet: 200 ImageNet-A images and 200 non-ImageNet ImageNet-Hard images drawn from the judge-audit samples (seed 1107), 20 blind repeats of earlier items at least 50 items apart (seed 1108), and 30 ImageNet validation images on whose truth both relabelings agree (15 correct and 15 wrong labels; 15 fine-grained and 15 coarse classes; seed 1110), in an order fixed by seeds 1109 and 1111. Each item showed the image with its benchmark label set (class names and dictionary definitions), and the annotator answered \emph{correct}, \emph{wrong} or \emph{unclear} under neutral written instructions equivalent to the judges' rule. No model output or judge call was shown, and no personal data was collected (no names, dates or demographic fields). The annotator participated voluntarily. No IRB review was carried out and no written consent was sought, because the task posed minimal risk (judging whether public benchmark images match their labels) and collected no personal data; the sheet contained only the task instructions. The analysis was settled in three dated amendments before any label existed. The annotator worked in a copy of the sheet; its 470 rows, 450 item numbers and instructions are identical to the original, and only the answer column was filled.

\paragraph{Answers and annotator quality.} All 450 items were answered, 126 (28\%) of them \emph{unclear}. On the known-truth controls, the annotator decided 29 of 30 and was right on 86\% (fine-grained 86\%, coarse 87\%); no correct label was called wrong (0\%), and 29\% of the wrong labels were called correct. On the 12 repeats decided both times, the two answers agree in 100\% of cases.

\paragraph{Estimates.} Both sources have all 200 items answered, and the annotator's label is the human reference. \cref{tab:supp_human} gives the share of wrong labels among decided images with bootstrap intervals, the share corrected for the measured annotator errors ($p=(p_{\rm obs}-f_w)/(1-f_w-f_c)$ with $f_w=$ 0\%, $f_c=$ 29\%), bounds that count every unclear answer as correct or as wrong, and the judges' joint call against the human label. ObjectNet's labels come from how its photos were taken, so a human ``wrong'' there may partly mean an object that cannot be recognized rather than a wrong label. Weighted by source size, about 36\% of ImageNet-Hard's images have a wrong label (ImageNet half by ReImageNet, other sources by the annotator; 44\% with the corrected share). We also compute a judge-calibrated estimate, $(q-\mathrm{FPR})/(\mathrm{TPR}-\mathrm{FPR})$ with the judges' rates measured against these labels; it gives 39\% (30--49\%) for ImageNet-Hard's other sources, close to the direct share, and an unstable 43\% (16--70\%) for ImageNet-A, where the judges flag 1 of 14 human-wrong labels. We report the direct human share. Dropping every item marked unclear on any occurrence changes nothing (10\% and 40\%). Recomputing the judges' rates with the matching annotator-error correction gives a true-positive rate of 0.65 and a false-positive rate of 0.00 for ImageNet-Hard's other sources (clipped at zero: the correction attributes all judge flags on human-``correct'' images to the annotator's misses) and a judge-calibrated estimate of 47\%; for ImageNet-A, the rates stay degenerate (0.07 and 0.01). The single annotator, the small controls (14 decided wrong labels) and the high share of unclear answers limit these estimates.

\begin{table*}[t]\centering\small\setlength{\tabcolsep}{4pt}
\begin{adjustbox}{max width=\linewidth}\begin{tabular}{@{}lrrccccc@{}}\toprule
Source & Decided & Unclear & Wrong (95\% CI) & Corrected & Bounds & Judge-calibrated & Judges P / R\\\midrule
ImageNet-A & 140 & 60 & 10\% (95\% CI 5--16\%) & 14\% & 7--37\% & 43\% (16--70\%)$^\dagger$ & 0.33 / 0.07\\
ImageNet-Hard, other & 143 & 57 & 41\% (95\% CI 33--48\%) & 57\% & 29--57\% & 39\% (30--49\%) & 0.84 / 0.65\\
\quad sketches & 96 & -- & 34\% (95\% CI 24--44\%) & 48\% & -- & -- & 0.81 / 0.67\\
\quad ObjectNet & 32 & -- & 66\% (95\% CI 50--81\%) & 92\% & -- & -- & 0.92 / 0.60\\
\bottomrule\end{tabular}\end{adjustbox}
\caption{\textbf{One annotator's blind labels.} Wrong: share of decided images with a wrong label; corrected: for the annotator's measured error rates on known-truth controls; bounds: unclear answers counted as correct or as wrong; judge-calibrated: the estimate $(q-\mathrm{FPR})/(\mathrm{TPR}-\mathrm{FPR})$ from the judges' flag rate $q$ on the whole audit sample and their rates against the human label; judges: precision and recall of both judges' joint call against the human label. $^\dagger$Unstable: the judges flag 1 of 14 human-wrong labels ($\mathrm{TPR}-\mathrm{FPR}=0.06$).}
\label{tab:supp_human}
\end{table*}

\section{Follow-up analyses}
\label{sec:supp_followups}
\begin{figure}[t]
\centering
\includegraphics[width=0.5\linewidth]{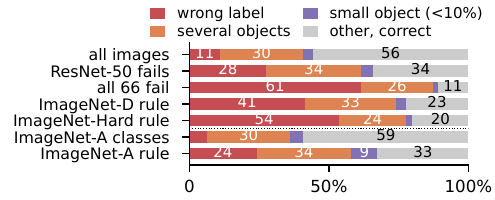}
\caption{\textbf{What a screen lets through} (ImageNet validation images, ReImageNet). Strict screens let through mostly wrong labels and multi-object scenes. ImageNet-A's screen, replayed on its 200 classes, also lets through correctly labeled images whose object covers less than a tenth of the frame. Follow-up analysis.}
\label{fig:composition}
\end{figure}

\begin{figure*}[t]
\centering
\includegraphics[width=\linewidth]{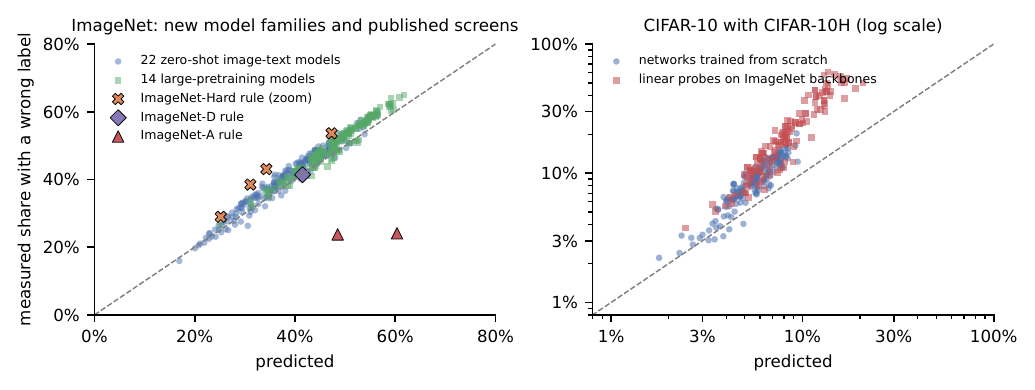}
\caption{\textbf{Predicted vs.\ measured shares of survivors with a wrong label.} Left: screens built from two model families that were not used in the fit, on ImageNet, and replays of published screens (ImageNet-Hard's rule at 1, 9, 36 and 324 crops; ImageNet-A's rule with and without its confidence condition). Right: CIFAR-10 screens from networks trained from scratch and from linear probes on ImageNet backbones; points above the diagonal are under-predicted.}
\label{fig:oos}
\end{figure*}

All analyses added after the held-out evaluations, with their result files: (1) composition of survivors by object count and size (Fig.~3; {\tt composition.json}); (2) ImageNet-A's ten-crop rule with eight filters in the 200- and 1000-class answer sets (\cref{tab:supp_ina}) and a factorial on fixed image IDs (\cref{tab:supp_factorial}) ({\tt posthoc\_P5.json}, {\tt posthoc\_review.json}); (3) in-sample center-view screens restricted to the 200 classes ({\tt posthoc\_labelspace.json}); (4) zoom rescue, overall (\cref{tab:supp_zoom}) and by object size ({\tt posthoc\_zoom.json}, {\tt posthoc\_review2.json}); (5) survival of wrongly and correctly labeled images under single CIFAR-10 and ImageNet filters ({\tt posthoc\_P6.json}); (6) a fitted constant-survival baseline, an image-level cross-fit, survivors under multi-label scoring and the gains of the models that proposed the judges' alternatives ({\tt posthoc\_review.json}); (7) bounds for survivors with an unsure annotation and the confusion between the two truths ({\tt posthoc\_review2.json}); (8) CIFAR-10 screens with at least 50 survivors, scored with the main baseline definition ({\tt posthoc\_strict.json}); (9) strict scoring of ImageNet-Hard's ImageNet half ({\tt posthoc\_strict.json}); (10) the comparison with a ranking by average label confidence ({\tt posthoc\_detector.json}); (11) the matched-accuracy CIFAR-10 comparison ({\tt posthoc\_P6.json}); (12) filter accuracy and the curve's errors ({\tt posthoc\_quality.json}); (13) leave-one-model-out refits ({\tt posthoc\_lomo.json}); (14) the judge correction on known truth ({\tt audit.json}); (15) ImageNet-A's rule replayed with all classes (Fig.~1b arrow; {\tt tool\_validation.json}); (16) the verified hard set ({\tt verified\_hard.json}; \cref{tab:supp_verified}); (17) the published VQA screens VQA-CE and AdVQA, scored with their own answers ({\tt vqa\_published.json}; Sec.~\ref{sec:supp_vqa}); (18) the full judge audit of ImageNet-Hard's other sources and the whole-benchmark corrected leaderboard ({\tt audit\_full.json}, {\tt leaderboard\_full.json}).

\paragraph{Why the ImageNet-A replay departs from the curve.} In our follow-up analyses, the 10{,}000 images of ImageNet-A's 200 classes point to the restricted answer set. With all 1000 classes allowed, the same ten-crop rule lets through 6.9\% of the images, of which 28\% have a wrong label, within 4.0 points of its prediction. With 200 classes it lets through 2.3\%, of which 24\% have a wrong label, and dropping the confidence condition changes this little (24\%). Likewise, restricting our first-family screens to the 200 classes lets three to eight times fewer images through, while the error rate barely changes (29\% with and 29\% without the restriction, for single filters). The restriction removes confusions with the other 800 classes. These confusions appear to be about as common for correctly as for wrongly labeled images, so the screen may look strict without being more selective for errors. The screen also lets through a different kind of correctly labeled image (\cref{fig:composition}). In these survivors, the labeled object covers a median of 0.21 of the image area (95\% CI 0.13--0.28), compared with 0.44 for all images of the 200 classes. By contrast, ImageNet-Hard's zoom rule lets almost no small-object images through.

\paragraph{Stronger filters.} We find that stronger filters let through more wrong labels at a given step. Across the 66 classifiers, the share of wrong labels among a model's own center-crop mistakes rises with its accuracy ($r=0.95$), from 18\% for AlexNet to 42\% for the most accurate model. However, since a model's error rate is exactly the share of images that survive its single-crop screen, this trend largely restates the curve. It suggests that a screen built from today's models may start where screens built from older models ended.

\paragraph{CIFAR-10 screens with at least 50 survivors.} Restricting the CIFAR-10 evaluation to screens with at least 50 survivors lowers the curve's error to 4.7 points, compared with 10.0 for the better baseline; for the linear probes alone, the error stays at 8.1 points (baseline 14.4).

\paragraph{VQA v2 examples.} Our random VQA v2 examples (Sec.~\ref{sec:supp_vqa}) include ambiguous and subjective questions.

\paragraph{Judge option ablation.} In an ablation on 150 screen-selected images (146 scored), the choice of options shown to the judges matters little: hiding the model-suggested alternatives moves the share called mislabeled from 38\% to 35\% and precision from 0.88 to 0.90.

\paragraph{Proposer models and the screen's own filter.} In a follow-up comparison, the three models that proposed the judges' alternatives gain slightly more than other top models (9.6 vs.\ 8.7 points on average). Meanwhile, CLIP ViT-L/14, which built the screen, moves from rank 103 to 99 of 103.

\paragraph{Strict scoring.} On the 4{,}767 of 5{,}464 decided images that contain at least one ImageNet class, BEiTv2-L goes from 64.3\% (original answers) to 65.3\% (strict answers), $+$1.0 point (95\% CI $-$0.2 to 2.2); Kendall's $\tau_b$ across the 103 models is 0.93, and top-ten models move by up to 2 places. An earlier version of this analysis compared strict accuracy on these 4{,}767 images with original accuracy on all 5{,}464 (62.5\%); about 1.7 of its 2.7 points came from dropping the 697 images without an ImageNet class, on which the model scored 50.8\% against the original labels. The paper reports only the like-for-like comparison.

\paragraph{Leave-one-model-out refits.} Dropping each of the 66 fitted models in turn, removing every in-sample screen that contains it and refitting the curve leaves the single-law factor between 1.39 and 1.40 per halving. The held-out errors stay between 2.3 and 2.4 points for the zero-shot family, 2.2 and 2.5 for the large-pretraining family, 5.5 and 7.3 for the ImageNet-Hard replay and 0.0 and 0.3 for the ImageNet-D replay. The paper uses the curve fitted on all 66 models.

\paragraph{Filter accuracy and the curve's errors.} On the 394 scored screens of the two held-out families, the difference between the measured and predicted error rate grows with the mean accuracy of the screen's filters (correlation 0.45; 1.6 to 2.9 points per ten points of accuracy). A linear accuracy correction fitted on one family lowers the other family's mean error from 2.4 to 1.7 points for the zero-shot models but raises it from 2.4 to 2.9 points for the large-pretraining models, so we keep the curve without an accuracy term.

\begin{table}[t]\centering\small\setlength{\tabcolsep}{2.5pt}
\caption{Follow-up analysis: ImageNet-A's ten-crop rule (low confidence on all crops) replayed with eight filters, in the 200-class label space and in the full 1000-class space. $n$: decided survivors; wrong labels in \%; area: median labelled-object area of correctly labeled survivors. This re-run of the ten-crop outputs differs from the replay in the main text by one image (223 vs.\ 224 survivors; 24.2\% vs.\ 24.1\%).}
\label{tab:supp_ina}
\begin{adjustbox}{max width=\linewidth}\begin{tabular}{lrrrrr}\toprule
 & \multicolumn{3}{c}{200 classes} & \multicolumn{2}{c}{1000 classes}\\
Filter & $n$ & Wrong & Area & $n$ & Wrong\\\midrule
tv\_alexnet & 861 & 14.1 & 0.23 & 1936 & 14.5 \\
tv\_resnet18 & 373 & 21.7 & 0.24 & 1010 & 23.1 \\
tv\_resnet50 & 223 & 24.2 & 0.20 & 676 & 28.1 \\
tv\_vit\_b\_16 & 167 & 33.5 & 0.20 & 601 & 34.9 \\
clip\_openai\_vitl14 & 160 & 31.2 & 0.20 & 747 & 32.9 \\
tv\_convnext\_large & 135 & 38.5 & 0.24 & 591 & 40.3 \\
hub\_dinov2\_vitg14\_lc & 82 & 41.5 & 0.17 & 371 & 45.0 \\
timm\_beitv2\_large\_22k & 58 & 56.9 & 0.22 & 349 & 54.2 \\
\bottomrule\end{tabular}\end{adjustbox}\end{table}
\begin{table}[t]\centering\small
\caption{Follow-up analysis: among images a model misclassifies at the center view, the share that at least one of the 324 zoom crops classifies as the original label (\%).}
\label{tab:supp_zoom}
\begin{adjustbox}{max width=\linewidth}\begin{tabular}{lrr}\toprule
Model & Valid & Mislabeled\\\midrule
tv\_resnet50 & 90.0 & 78.3 \\
tv\_resnet18 & 87.4 & 75.6 \\
tv\_alexnet & 80.1 & 68.4 \\
tv\_vgg16\_bn & 88.0 & 75.6 \\
tv\_vit\_b\_32 & 92.7 & 78.6 \\
clip\_openai\_vitl14 & 91.1 & 74.9 \\
\bottomrule\end{tabular}\end{adjustbox}\end{table}

\begin{table*}[t]\centering\small
\caption{Follow-up analysis: ImageNet-A's rule on fixed image IDs. ResNet-50 (torchvision) with ten crops must misclassify all crops; the confidence condition additionally requires a probability of at most 0.15 for the label on all crops. Rows 1--4 use the 10{,}000 validation images of ImageNet-A's 200 classes. Rates in \%. Row 1 differs from the replay in the main text by one image (223 vs.\ 224 survivors).}
\label{tab:supp_factorial}
\begin{adjustbox}{max width=\linewidth}\begin{tabular}{lrrrrr}\toprule
Setting & Decided pool & $\pi$ & Survivors & $f$ & $\theta$\\\midrule
200 classes, confidence $\le 0.15$ (published rule) & 9{,}766 & 6.5 & 223 & 2.3 & 24.2 \\
200 classes, no confidence condition & 9{,}766 & 6.5 & 327 & 3.3 & 23.9 \\
1000 classes, confidence $\le 0.15$ & 9{,}766 & 6.5 & 676 & 6.9 & 28.1 \\
1000 classes, no confidence condition & 9{,}766 & 6.5 & 989 & 10.1 & 28.0 \\
All 1000 classes, center view only & 48{,}912 & 11.1 & 11{,}518 & 23.5 & 27.6 \\
\bottomrule\end{tabular}\end{adjustbox}\end{table*}

\begin{table*}[t]
\centering
\small
\setlength{\tabcolsep}{4pt}
\caption{\textbf{Consequences of label corrections} for 103 models. Wrong: share of images without a correct accepted answer (ReImageNet) or called mislabeled by both judges (\%). Best model: highest accuracy (\%) with the original answers and with lenient scoring (corrected classes added to the accepted answers, original label kept). Strict (follow-up analysis): accuracy of the same model with the original answers and when only classes that ReImageNet found in the image count, both on the 4{,}767 images that contain at least one ImageNet class. For ImageNet-A, images whose corrected class lies outside its 200 classes are dropped (17 of 24). Kendall $\tau$ and filter rank use lenient scoring. Top-5 errors: share of the five most accurate models' errors that fall on images with a wrong or questioned label. Filter rank: rank of the model that built the screen (CLIP ViT-L/14 for ImageNet-Hard, ResNet-50 for ImageNet-A) among 103. Judges: the joint call of both judges; against one annotator's blind labels it finds about two thirds of the wrong labels in ImageNet-Hard's other sources and almost none in ImageNet-A, so the judge rows likely understate the consequences.}
\label{tab:supp_cons}
\begin{adjustbox}{max width=\linewidth}\begin{tabular}{llrrccccc}
\toprule
Benchmark & Truth & $n$ & Wrong & Best model & Strict & Top-5 errors & Kendall $\tau$ & Filter rank \\
\midrule
ImageNet-Hard, ImageNet half & ReImageNet & 5{,}464 & 32.4 & 62.5 $\to$ 70.5 & 64.3 $\to$ 65.3 & 47.2--51.4 & 0.95 & 103 $\to$ 101 \\
ImageNet-Hard, sample & judges & 598 & 23.2 & 42.5 $\to$ 53.0 & -- & 31.4--36.8 & 0.90 & 103 $\to$ 99 \\
ImageNet-A, sample & judges & 600 & 4.0 & 91.5 $\to$ 91.9 & -- & 5.4--10.2 & 0.99 & 103 $\to$ 103 \\
\bottomrule
\end{tabular}\end{adjustbox}
\end{table*}

\begin{table*}[t]
\centering
\small
\caption{\textbf{Screen, then verify} (follow-up resource; the screen uses only the in-sample models, and only held-out models are scored). Screened: ImageNet validation images that at least $k$ of the 66 fitted classifiers get wrong; wrong: share with a wrong label (ReImageNet, decided images); verified: screened images that both relabelings confirm as correctly labeled single-object images. Accuracies (\%) are means over the 36 held-out models (maximum in parentheses); errors on wrong labels: share of their errors on the screened images that fall on wrong labels; $\tau$ and shift compare the models' rankings on the screened and verified sets. All validation images: mean accuracy 79\%.}
\label{tab:supp_verified}
\begin{adjustbox}{max width=\linewidth}\begin{tabular}{lrrrrrrrr}
\toprule
Screen & Screened & Wrong & Accuracy & Errors on & Verified & Accuracy & $\tau$ & Largest \\
 & & (\%) & (screened) & wrong labels (\%) & & (verified) & & rank shift \\
\midrule
40 of 66 & 7{,}303 & 38 & 28 (41) & 45 & 694 & 50 (74) & 0.89 & 5 \\
50 of 66 & 5{,}735 & 42 & 22 (35) & 48 & 444 & 44 (70) & 0.85 & 8 \\
66 of 66 & 1{,}892 & 61 & 7 (13) & 63 & 41 & 25 (49) & 0.72 & 13 \\
\bottomrule
\end{tabular}\end{adjustbox}
\end{table*}

\paragraph{Screen, then verify.} \cref{tab:supp_verified} gives the verified hard set at three screen strengths. The construction uses only the in-sample models: images that at least three quarters of the 66 fitted classifiers get wrong, kept when ReImageNet (decided, label valid, single object) and ReaL (exactly the original class) both confirm the label. Only the 36 held-out models, which played no part in the screen, are scored. The index lists the screened images (our model outputs); {\tt code/tool/verify\_hard.py} rebuilds the verified set from the user's own ReImageNet and ReaL copies, so no label is redistributed.

\subsection{Full ImageNet-Hard audit: judge-only measurements (follow-up)}
This subsection reports full-audit judge-only measurements. The main text's headline estimates are the human-label and judge-calibrated ones (Sec.~6), and its sample-based figures are unchanged.

\paragraph{Full audit of ImageNet-Hard's other sources.} After the sample results, we ran the same judge protocol on all 5{,}090 non-ImageNet images of ImageNet-Hard ({\tt design/INH\_FULL\_AUDIT.md}; the 277 sampled images reuse their calls). Both judges answered 5{,}089 images (one image received no valid answer from one judge after six attempts). Both call 35.6\% (95\% CI 34.3--36.9) mislabeled: 36\% of the sketches, 36\% of the ObjectNet images and 28\% of the three smaller sources; 22\% are undecided. Corrected with the judges' error rates on the four ImageNet image sets, these judge-only measurements give 42\% to 87\%; corrected with their rates measured against the human labels, they give 48\%. The direct human-label share (41\%) and the judge-calibrated estimate on the annotated sample (39\%) remain the headline estimates. Weighted by source size with the directly measured ImageNet half, the whole benchmark lies between 37\% and 58\% by the judges (40\% with the human-measured rates), compared with 36\% from the human labels. The sample's values ({\tt audit.json}) lie within these intervals or near their edges.

\paragraph{Judge-corrected scores on the 598-image sample and on ImageNet-A.} Because the annotator only says whether a label is right, for the full 598-image ImageNet-Hard sample we use lenient scoring with the class that both judges name for each image they call mislabeled. The best model's accuracy rises from 42.5\% to 53.0\% (10.5 points, 95\% CI 8.0--13.0), and for the five most accurate models, 31\% to 37\% of the original errors fall on images that the judges call mislabeled. Since our judges miss about a third of the wrong labels that our annotator finds, these numbers may understate the changes. On ImageNet-A, lenient scoring with the judges' classes moves the best model's accuracy by only 0.4 points ($\tau=0.99$), since the judges miss almost all errors that our annotator finds there; images whose corrected class lies outside its 200 classes are dropped (17 of 24).

\paragraph{Whole-benchmark corrected leaderboard.} All 103 models were scored ({\tt design/LEADERBOARD\_FULL.md}) on all 10{,}980 images with lenient corrected answers: ReImageNet's classes added for the 5{,}464 decided matched ImageNet-sourced images, the judges' agreed class added for the non-ImageNet images they both call mislabeled, and the accepted labels unchanged for the remaining 427 images. The best model goes from 42.2\% to 55.5\% (13.3 points, 95\% CI 12.7--14.0); Kendall's $\tau$ is 0.90 (95\% CI 0.89--0.91); top-ten models move by at most 1 place; CLIP ViT-L/14, which built the screen, moves from rank 103 to 98. For the five most accurate models, 42\% to 48\% of the original errors fall on images flagged as mislabeled (by ReImageNet for the ImageNet half, by both judges elsewhere). Per-model accuracies are in {\tt leaderboard\_full.json}; Table~S10 keeps the sample-based values.

\subsection{ImageNet-V2: a new pool}
\label{sec:supp_v2}
\paragraph{Setup.} We set the screens, the predictor and the judging sample before running any model on ImageNet-V2 ({\tt design/V2\_PROSPECTIVE.md}).
\begin{itemize}
\item \emph{Pool.} ImageNet-V2 MatchedFrequency~\citep{recht2019imagenetv2}: 10{,}000 images, ten per class.
\item \emph{Screens.} All screens use the center view and the top-1 prediction against ImageNet-V2's own label. In S1, one ResNet-50 fails; in S2, at least 50 of our 66 fitted classifiers fail; in S3, all 66 fail.
\item \emph{Models.} All 103 roster models were run with their roster preprocessing, checkpoints and software versions. On 781 ImageNet validation images, the rerun matched the stored outputs exactly for three models and agreed on 99.6\% of top-1 predictions for ResNet-50, where three near-ties flipped.
\item \emph{Predictions.} The curve of each screen's type predicts the share of wrong labels from the screen's survival share and the pool's error rate, which is measured with the truth below; the predictions are therefore conditional on that rate.
\end{itemize}

\paragraph{Truth.} The primary truth is the human relabeling of \citet{anzaku2024multilabelfy}, which gives refined multi-labels for all 10{,}000 images. Annotators accepted or rejected classes proposed by models, and experienced annotators then refined the result. This follows ReaL's protocol, so predictions use the curve fitted under ReaL. A label counts as wrong when its class, together with equivalent duplicate classes, is not among the refined labels. The secondary truth is our two judges' joint call on a sample of 2{,}437 images, corrected with their ImageNet rates (random-image rates for a random 400 images of the pool, screen-selected rates for survivors); those predictions use the curve fitted under ReImageNet.

\paragraph{Results.} \cref{tab:supp_v2} gives the screens. Under the human relabeling, the mean error is 10.3 points (95\% CI 8.7--11.7) against 17.9 for the better baseline: the curve is accurate for the mildest screen and too low for the strict ones. With the judges as truth, the mean error is 11.2 points (95\% CI 6.1--31.8) against 20.7, and the predictions for the strict screens are too high (S3: 53\% measured, 58\% predicted). On the images that both judges decided, the judges flag 414 of the 721 images that the relabeling calls mislabeled (recall 0.57), and their flags are right for 0.74 of flagged images.

\paragraph{Verified subset.} Of the 2{,}061 survivors of S2, the relabeling verifies 265 as showing only their labeled class. Our 36 held-out models reach 23\% accuracy on all S2 survivors and 45\% on the verified ones, against 72\% on the whole pool. Among their S2 errors, 49\% fall on wrong labels.

\paragraph{Overlap with ImageNet-Hard.} ImageNet-Hard drew 81 of its images from ImageNet-V2 through its CLIP screen, and our whole-benchmark judge audit (\cref{sec:supp_followups}) had judged those 81 images with ImageNet-Hard's labels. The predictions use neither those calls nor any relabeling, and the judging sample was judged afresh.

\begin{table*}[t]
\centering
\small
\caption{\textbf{ImageNet-V2 screens.} Survival share $f$ and error rates in \%. Human: the relabeling of \citet{anzaku2024multilabelfy}, predictions from the curve fitted under ReaL. Judges: joint call of the two judges on a sample of the screens (random 300 of S1, a random 2{,}000 of S2 including all of S3, all of S3, and a random 400 of the pool), corrected with the ImageNet judge rates, predictions from the curve fitted under ReImageNet.}
\label{tab:supp_v2}
\begin{adjustbox}{max width=\linewidth}\begin{tabular}{lrrrrrrrrr}
\toprule
 & & \multicolumn{4}{c}{Human relabeling (primary)} & \multicolumn{4}{c}{Two judges (secondary)} \\
\cmidrule(lr){3-6}\cmidrule(lr){7-10}
Screen & $f$ & $n$ & measured & predicted & error & $n$ & measured & predicted & error \\
\midrule
S1: ResNet-50 fails & 36.8 & 3{,}681 & 27.1 & 26.5 & 0.6 & 300 & 5.5 & 25.0 & 19.5 \\
S2: $\geq$50 of 66 fail & 20.6 & 2{,}061 & 42.8 & 33.5 & 9.3 & 2{,}000 & 27.5 & 36.7 & 9.2 \\
S3: all 66 fail & 7.3 & 726 & 66.7 & 45.7 & 20.9 & 726 & 53.2 & 58.2 & 5.0 \\
Pool (all images) & 100.0 & 10{,}000 & 11.9 & -- & -- & 400 & 13.1 & -- & -- \\
\midrule
Mean error (95\% CI) / better baseline & \multicolumn{5}{c}{10.3 [8.7, 11.7] / 17.9: fail} & \multicolumn{4}{c}{11.2 [6.1, 31.8] / 20.7: fail} \\
\bottomrule
\end{tabular}\end{adjustbox}
\end{table*}

\subsection{Effective robustness under corrected labels}
\label{sec:supp_effrob}
\paragraph{Setup.} We set this analysis before computing any effective robustness ({\tt design/EFFROB\_CORRECTED.md}).
\begin{itemize}
\item \emph{Line.} For each labeling of ImageNet-Hard, we fitted a line in logit space to the 67 supervised classifiers, relating benchmark accuracy to ImageNet validation accuracy. ImageNet accuracy always uses the original labels.
\item \emph{Labelings.} The original answers, and the whole-benchmark corrected answers. The corrected answers come from ReImageNet's classes on the ImageNet half and from the class that both judges name on the other sources (\cref{sec:supp_followups}).
\item \emph{Effective robustness.} A model's benchmark accuracy minus the line's prediction~\citep{taori2020shifts}.
\item \emph{Intervals.} Paired bootstrap over 1{,}000 resamples of ImageNet-Hard images, refitting the line in each.
\end{itemize}

\paragraph{Results.} For the 22 zero-shot image--text models, the mean effective robustness is $-$3.7 points (95\% CI $-$4.2 to $-$3.2) with the original answers and $+$2.1 (95\% CI 1.5--2.7) with the corrected answers; the difference is $+$5.8 (95\% CI 5.4--6.1), so the sign of the reading changes. For the 14 large-pretraining models, the mean goes from $+$1.3 to $+$2.8, so their reading keeps its sign. Kendall's $\tau$ between the models' effective robustness under the two labelings is 0.51. The largest single change is a LAION-trained CLIP ViT-L/14, from $-$5.2 to $+$1.3.

\paragraph{Follow-up analyses.} On the ImageNet half alone, which has human corrections and no judges, the zero-shot mean rises from $-$12.5 to $-$8.4 points without changing sign. On the other sources, which have judge corrections whose candidate classes were proposed in part by an image--text model, it rises from $+$6.5 to $+$14.7. Without the three proposer models, the whole-benchmark change is from $-$3.9 to $+$1.9. Correction raises the zero-shot models' effective robustness on both halves, and the sign reverses only in the whole-benchmark mean. A possible reason, which we have not tested, is that ImageNet-Hard's screen used CLIP itself, so its wrong labels may cost image--text models the most.

\section{Datasets and licences}
\label{sec:supp_data}
ImageNet's validation images were used under the ImageNet terms of access and are not redistributed. VQA-CE's question-ID lists (GitHub release of its authors; no licence stated) and AdVQA's validation annotations (CC BY-NC 4.0) were used for computation only and are not redistributed; scripts download them. ReaL labels are released under Apache-2.0. ReImageNet labels are released under CC BY 4.0 with access terms that restrict use to research, require citation and forbid redistribution, so the code and data will include only aggregates and a script that rebuilds per-image results from the gated release. CIFAR-10 was used under its original terms, and CIFAR-10H is released under CC BY-NC-SA 4.0. ImageNet-A and ImageNet-Hard are released under the MIT licence, and ImageNet-Hard's images keep the terms of their source datasets: ImageNet-Sketch, ImageNet-R and ImageNet-V2 were released by their authors for research use, and ObjectNet's licence forbids using it to train or tune models and posting its images without the border; no images from any of them are redistributed. VQA v2 annotations are released under CC BY 4.0, and its COCO val2014 images keep the COCO and Flickr terms; only per-question flags and model answers will be included. We redistribute no third-party labels or annotations: our scripts download them from the official sources. All model weights were used for evaluation only, under their respective licences, some of which restrict commercial use.

\section{VQA v2 evaluation}
\label{sec:supp_vqa}
The design is in \texttt{design/VQA\_EXTENSION.md}, with one amendment (\texttt{design/VQA\_AMENDMENT\_1.md}). The scout sample (2{{,}}000 questions) checked each model's answer format; models below 50\% VQA accuracy were dropped{,} and three replacement models were added before any screen statistic was computed. A third replacement model was run on the scout questions and dropped by the same rule; it was later found ineligible under the study's model policy{,} its outputs were deleted{,} and no code or result for it will be released. The confirmatory sample has 50{,}000 questions{,} disjoint from the scout. \begin{table*}[t]\centering\small\setlength{\tabcolsep}{5pt}\begin{adjustbox}{max width=\linewidth}\begin{tabular}{@{}llrr@{}}\toprule Model & Language model; vision encoder & Scout acc. & Full acc. \\\midrule
Pixtral-12B & Mistral NeMo; Pixtral encoder & 78.9 & 78.8 \\
Phi-3.5-vision & Phi-3.5-mini; CLIP ViT-L/14 & 73.8 & 74.2 \\
LLaVA-1.6-Mistral-7B & Mistral-7B; CLIP ViT-L/14-336 & 81.3 & 80.2 \\
Mistral-Small-3.1-24B & Mistral Small 3.1; own encoder & 79.8 & 79.1 \\
LLaVA-1.5-7B & Vicuna-7B (Llama 2); CLIP ViT-L/14-336 & 75.1 & 75.8 \\
LLaVA-1.5-13B & Vicuna-13B (Llama 2); CLIP ViT-L/14-336 & 77.5 & 76.9 \\
Llama-3.2-11B-Vision & Llama 3.1; own encoder & 44.9 & dropped (format rule) \\
Molmo-7B-O & OLMo-7B; CLIP ViT-L/14-336 & 43.3 & dropped (format rule) \\
\bottomrule\end{tabular}\end{adjustbox}\caption{\textbf{VQA v2 models.} VQA accuracy (\%) on the 2{,}000-question scout and the 50{,}000-question confirmatory sample. Models below 50\% on the scout were dropped by the format rule.}\label{tab:supp_vqa}\end{table*}
\paragraph{Disputed answers by question type.} Among all questions, 0.2\% of yes/no, 20\% of number and 34\% of other questions have a disputed answer. Among the questions that all models fail, the shares are 52\% (number), 79\% (other), 1\% (yes/no). With the stricter threshold (at most 3 of 10 annotators gave the official answer), 13\% of all questions and 53\% of the questions that all models fail are disputed.
\paragraph{Examples.} Questions that all models fail and whose official answer is disputed (drawn at random):
\begin{itemize}\small
\item \emph{What is this man doing?} Official answer: ``dissecting worm'' (given by 2 of 10). Annotators: dissecting worm, cutting specimen, fixing something, science, dissecting, looking at something, watching, examining, looking at something, dissecting worm. Models: Craft, Examining specimen, Looking, Looking at computer, Standing.
\item \emph{What kind of traveler would appreciate the advertisement on the bus?} Official answer: ``london traveler'' (given by 3 of 10). Annotators: international, adventurer, tourist, london traveler, london traveler, british, bus traveler, bus, tourist, london traveler. Models: Tourist.
\item \emph{Why is the stripe this way on this animal?} Official answer: ``photoshop'' (given by 1 of 10). Annotators: heredity, breed, mixed breed, painted, hybrid, photoshop, zebra, birth defect, zebra horse, born that way. Models: Camouflage, Genetics, Zebra.
\end{itemize}

\paragraph{Published VQA screens.} We also applied the curve to two published VQA screens. VQA-CE~\citep{dancette2021vqace} mines shortcut rules on VQA v2's training questions and calls a validation question a counterexample when at least one rule matches it and none of the matching rules' answers is among its ten human answers; its 63{,}298 counterexamples are 29.5\% of the 214{,}354 validation questions. We treat the rule model as a single filter (screen type \emph{one}); the prediction is 41.5\% (type \emph{all-of-n}: 47.3\%), and the baselines are 19.2\% and $\min(1,\pi/f)$. Measured: 40.2\% (95\% CI 39.8--40.6), an error of 1.3 points against 21.0 for the better baseline. VQA-CE's ``hard'' questions, which no rule matches, have 34.7\% disputed answers and its remaining questions 9.9\%. VQA-CE counts a failure only when no annotator gave the rule's answer, unlike our screens, which score against the official answer. AdVQA~\citep{sheng2021advqa} has no single official answer, so both sets are compared on the share of questions whose most common normalized answer was given by at most four of ten annotators (for VQA v2 this agrees with the official-answer definition on all but 14 questions). We expected AdVQA, which kept questions that fooled models, to have a higher share than VQA v2, but it has a lower one: 10.5\% vs.\ 19.2\% (difference 95\% CI -9.3 to -8.1 points), also within answer types (``other'': 19.7\% vs.\ 33.4\%). AdVQA's builders showed each question with the model's answer to two further annotators (a third broke ties), who judged whether the model's answer was correct and whether the question was valid, i.e., answerable from the image; a question was kept only if at least two judged it valid. They report higher agreement among annotators than on VQA v2; our reading that this validation removed ambiguous questions was made after the result. AdVQA's questions were also written by people and its images differ (COCO 2017 validation).

\section{ScreenAudit: tool and per-image files}
\label{sec:supp_tool}
ScreenAudit consists of {\tt code/tool/screen\_audit.py} with the fitted curve ({\tt curve.json}), a validation script and tests. In replay mode, the tool takes a screen's survivors on ImageNet's validation images (or the filters' predictions and a rule) and reports the share of survivors, the share with a wrong label under ReImageNet (from the user's gated copy) and ReaL with bootstrap intervals, and the curve's prediction for each type of screen. With {\tt --label-space} below 1000 it warns that a restricted answer set makes the prediction unreliable and, given {\tt --full-mask} (the survivors of the same screen replayed with all classes), reports that replay as well; {\tt --pool} restricts the image set. In forecast mode, it takes only the share surviving, the type of screen and a base error rate, and returns the prediction with the range of $\pm$4.0 points described in the main text. The validation script reproduces the replay numbers of the main text (errors of 0.1, 6.3, 36.3 and 4.0 points for ImageNet-D, ImageNet-Hard, ImageNet-A with 200 classes and with all classes). It includes no third-party data. The per-image files ({\tt results/release}) hold, for each of the 5{,}646 matched ImageNet-Hard images of ImageNet origin, its position and parquet row, its matched validation file, the copy type and whether ReaL's raters accepted any of its accepted classes; the judges' calls for the 1{,}200 sampled images (1{,}198 judged) and, as a follow-up, for all 5{,}090 non-ImageNet ImageNet-Hard images (5{,}089 judged), 6{,}012 distinct images in all; and, for VQA v2, the disputed-answer flags of all 214{,}354 validation questions and the six models' answers on the 50{,}000-question sample. Scripts download ImageNet-Hard's accepted classes, ReaL's classes and the VQA v2 annotations from their official sources, under their terms, and rejoin or recompute them (a test checks that the rebuilt tables match ours exactly); a further script rebuilds each image's corrected answer set and status from the gated ReImageNet release (on our copy: 5{,}464 decided images, 1{,}772 without a correct label, 4{,}767 with a non-empty corrected answer set). Code and data (ScreenAudit and the per-image files) will be released later.

\section{Limitations in detail}
\label{sec:supp_limits}
All filters see the same random views, so the depth screens share their attempts. The ImageNet-Hard replay scores against the original label, while the benchmark scored against a label set (\cref{sec:supp_judges}). The human labels come from a single annotator, who marked 28\% of items unclear and whose accuracy on known-truth items is 86\%; the judges' recall against these labels is 0.65 on ImageNet-Hard's other sources but 0.07 on ImageNet-A. The judges are closed, versioned commercial models whose behavior may change. The VQA screens use exact match to the official answer rather than VQA's own metric, which credits any answer that at least three annotators gave; the stricter rule in the main text is closer to that metric. The held-out families test new filters on the same images rather than new images.

\section{Model roster}
\label{sec:supp_roster}
\begin{small}
\setlength{\tabcolsep}{3pt}
\begin{longtable}{p{3.4cm}c>{\raggedright\arraybackslash}p{4.3cm}>{\raggedright\arraybackslash}p{2.9cm}>{\raggedright\arraybackslash}p{2.6cm}r}
\caption{Model roster: 103 classifiers. Family A: supervised (in-sample); B: zero-shot image--text (held out); C: large-scale pretraining with an ImageNet head (held out). Last column: measured top-1 accuracy (\%) on ImageNet validation images (center view).}\label{tab:roster}\\
\toprule
Model & Fam. & Checkpoint & Released by & Training data & Top-1 \\
\midrule
\endfirsthead
\toprule
Model & Fam. & Checkpoint & Released by & Training data & Top-1 \\
\midrule
\endhead
tv\_\allowbreak{}alexnet & A & \texttt{\scriptsize alexnet /\allowbreak{} IMAGENET1K\_\allowbreak{}V1} & PyTorch/torchvision & IN-1k & 56.5 \\
tv\_\allowbreak{}vgg11\_\allowbreak{}bn & A & \texttt{\scriptsize vgg11\_\allowbreak{}bn /\allowbreak{} IMAGENET1K\_\allowbreak{}V1} & PyTorch/torchvision & IN-1k & 70.4 \\
tv\_\allowbreak{}vgg16\_\allowbreak{}bn & A & \texttt{\scriptsize vgg16\_\allowbreak{}bn /\allowbreak{} IMAGENET1K\_\allowbreak{}V1} & PyTorch/torchvision & IN-1k & 73.4 \\
tv\_\allowbreak{}vgg19\_\allowbreak{}bn & A & \texttt{\scriptsize vgg19\_\allowbreak{}bn /\allowbreak{} IMAGENET1K\_\allowbreak{}V1} & PyTorch/torchvision & IN-1k & 74.2 \\
tv\_\allowbreak{}squeezenet1\_\allowbreak{}1 & A & \texttt{\scriptsize squeezenet1\_\allowbreak{}1 /\allowbreak{} IMAGENET1K\_\allowbreak{}V1} & PyTorch/torchvision & IN-1k & 58.2 \\
tv\_\allowbreak{}resnet18 & A & \texttt{\scriptsize resnet18 /\allowbreak{} IMAGENET1K\_\allowbreak{}V1} & PyTorch/torchvision & IN-1k & 69.7 \\
tv\_\allowbreak{}resnet34 & A & \texttt{\scriptsize resnet34 /\allowbreak{} IMAGENET1K\_\allowbreak{}V1} & PyTorch/torchvision & IN-1k & 73.3 \\
tv\_\allowbreak{}resnet50 & A & \texttt{\scriptsize resnet50 /\allowbreak{} IMAGENET1K\_\allowbreak{}V1} & PyTorch/torchvision & IN-1k & 76.1 \\
tv\_\allowbreak{}resnet50\_\allowbreak{}v2 & A & \texttt{\scriptsize resnet50 /\allowbreak{} IMAGENET1K\_\allowbreak{}V2} & PyTorch/torchvision & IN-1k & 80.8 \\
tv\_\allowbreak{}resnet101\_\allowbreak{}v2 & A & \texttt{\scriptsize resnet101 /\allowbreak{} IMAGENET1K\_\allowbreak{}V2} & PyTorch/torchvision & IN-1k & 81.9 \\
tv\_\allowbreak{}resnet152\_\allowbreak{}v2 & A & \texttt{\scriptsize resnet152 /\allowbreak{} IMAGENET1K\_\allowbreak{}V2} & PyTorch/torchvision & IN-1k & 82.3 \\
tv\_\allowbreak{}resnext50\_\allowbreak{}v2 & A & \texttt{\scriptsize resnext50\_\allowbreak{}32x4d /\allowbreak{} IMAGENET1K\_\allowbreak{}V2} & PyTorch/torchvision & IN-1k & 81.2 \\
tv\_\allowbreak{}resnext101\_\allowbreak{}64x4d & A & \texttt{\scriptsize resnext101\_\allowbreak{}64x4d /\allowbreak{} IMAGENET1K\_\allowbreak{}V1} & PyTorch/torchvision & IN-1k & 83.2 \\
tv\_\allowbreak{}wide\_\allowbreak{}resnet50\_\allowbreak{}2\_\allowbreak{}v2 & A & \texttt{\scriptsize wide\_\allowbreak{}resnet50\_\allowbreak{}2 /\allowbreak{} IMAGENET1K\_\allowbreak{}V2} & PyTorch/torchvision & IN-1k & 81.6 \\
tv\_\allowbreak{}wide\_\allowbreak{}resnet101\_\allowbreak{}2\_\allowbreak{}v2 & A & \texttt{\scriptsize wide\_\allowbreak{}resnet101\_\allowbreak{}2 /\allowbreak{} IMAGENET1K\_\allowbreak{}V2} & PyTorch/torchvision & IN-1k & 82.5 \\
tv\_\allowbreak{}densenet121 & A & \texttt{\scriptsize densenet121 /\allowbreak{} IMAGENET1K\_\allowbreak{}V1} & PyTorch/torchvision & IN-1k & 74.4 \\
tv\_\allowbreak{}densenet201 & A & \texttt{\scriptsize densenet201 /\allowbreak{} IMAGENET1K\_\allowbreak{}V1} & PyTorch/torchvision & IN-1k & 76.9 \\
tv\_\allowbreak{}mobilenet\_\allowbreak{}v2\_\allowbreak{}v2 & A & \texttt{\scriptsize mobilenet\_\allowbreak{}v2 /\allowbreak{} IMAGENET1K\_\allowbreak{}V2} & PyTorch/torchvision & IN-1k & 72.1 \\
tv\_\allowbreak{}mobilenet\_\allowbreak{}v3\_\allowbreak{}large\_\allowbreak{}v2 & A & \texttt{\scriptsize mobilenet\_\allowbreak{}v3\_\allowbreak{}large /\allowbreak{} IMAGENET1K\_\allowbreak{}V2} & PyTorch/torchvision & IN-1k & 75.3 \\
tv\_\allowbreak{}mnasnet1\_\allowbreak{}0 & A & \texttt{\scriptsize mnasnet1\_\allowbreak{}0 /\allowbreak{} IMAGENET1K\_\allowbreak{}V1} & PyTorch/torchvision & IN-1k & 73.4 \\
tv\_\allowbreak{}regnet\_\allowbreak{}x\_\allowbreak{}1\_\allowbreak{}6gf\_\allowbreak{}v2 & A & \texttt{\scriptsize regnet\_\allowbreak{}x\_\allowbreak{}1\_\allowbreak{}6gf /\allowbreak{} IMAGENET1K\_\allowbreak{}V2} & PyTorch/torchvision & IN-1k & 79.6 \\
tv\_\allowbreak{}regnet\_\allowbreak{}y\_\allowbreak{}3\_\allowbreak{}2gf\_\allowbreak{}v2 & A & \texttt{\scriptsize regnet\_\allowbreak{}y\_\allowbreak{}3\_\allowbreak{}2gf /\allowbreak{} IMAGENET1K\_\allowbreak{}V2} & PyTorch/torchvision & IN-1k & 82.0 \\
tv\_\allowbreak{}regnet\_\allowbreak{}y\_\allowbreak{}16gf\_\allowbreak{}v2 & A & \texttt{\scriptsize regnet\_\allowbreak{}y\_\allowbreak{}16gf /\allowbreak{} IMAGENET1K\_\allowbreak{}V2} & PyTorch/torchvision & IN-1k & 82.9 \\
tv\_\allowbreak{}regnet\_\allowbreak{}y\_\allowbreak{}32gf\_\allowbreak{}v2 & A & \texttt{\scriptsize regnet\_\allowbreak{}y\_\allowbreak{}32gf /\allowbreak{} IMAGENET1K\_\allowbreak{}V2} & PyTorch/torchvision & IN-1k & 83.4 \\
tv\_\allowbreak{}vit\_\allowbreak{}b\_\allowbreak{}16 & A & \texttt{\scriptsize vit\_\allowbreak{}b\_\allowbreak{}16 /\allowbreak{} IMAGENET1K\_\allowbreak{}V1} & PyTorch/torchvision & IN-1k & 81.0 \\
tv\_\allowbreak{}vit\_\allowbreak{}b\_\allowbreak{}32 & A & \texttt{\scriptsize vit\_\allowbreak{}b\_\allowbreak{}32 /\allowbreak{} IMAGENET1K\_\allowbreak{}V1} & PyTorch/torchvision & IN-1k & 75.8 \\
tv\_\allowbreak{}vit\_\allowbreak{}l\_\allowbreak{}16 & A & \texttt{\scriptsize vit\_\allowbreak{}l\_\allowbreak{}16 /\allowbreak{} IMAGENET1K\_\allowbreak{}V1} & PyTorch/torchvision & IN-1k & 79.7 \\
tv\_\allowbreak{}vit\_\allowbreak{}l\_\allowbreak{}32 & A & \texttt{\scriptsize vit\_\allowbreak{}l\_\allowbreak{}32 /\allowbreak{} IMAGENET1K\_\allowbreak{}V1} & PyTorch/torchvision & IN-1k & 77.0 \\
tv\_\allowbreak{}swin\_\allowbreak{}t & A & \texttt{\scriptsize swin\_\allowbreak{}t /\allowbreak{} IMAGENET1K\_\allowbreak{}V1} & PyTorch/torchvision & IN-1k & 81.4 \\
tv\_\allowbreak{}swin\_\allowbreak{}s & A & \texttt{\scriptsize swin\_\allowbreak{}s /\allowbreak{} IMAGENET1K\_\allowbreak{}V1} & PyTorch/torchvision & IN-1k & 83.2 \\
tv\_\allowbreak{}swin\_\allowbreak{}b & A & \texttt{\scriptsize swin\_\allowbreak{}b /\allowbreak{} IMAGENET1K\_\allowbreak{}V1} & PyTorch/torchvision & IN-1k & 83.6 \\
tv\_\allowbreak{}swin\_\allowbreak{}v2\_\allowbreak{}t & A & \texttt{\scriptsize swin\_\allowbreak{}v2\_\allowbreak{}t /\allowbreak{} IMAGENET1K\_\allowbreak{}V1} & PyTorch/torchvision & IN-1k & 82.1 \\
tv\_\allowbreak{}swin\_\allowbreak{}v2\_\allowbreak{}b & A & \texttt{\scriptsize swin\_\allowbreak{}v2\_\allowbreak{}b /\allowbreak{} IMAGENET1K\_\allowbreak{}V1} & PyTorch/torchvision & IN-1k & 84.1 \\
tv\_\allowbreak{}convnext\_\allowbreak{}tiny & A & \texttt{\scriptsize convnext\_\allowbreak{}tiny /\allowbreak{} IMAGENET1K\_\allowbreak{}V1} & PyTorch/torchvision & IN-1k & 82.5 \\
tv\_\allowbreak{}convnext\_\allowbreak{}small & A & \texttt{\scriptsize convnext\_\allowbreak{}small /\allowbreak{} IMAGENET1K\_\allowbreak{}V1} & PyTorch/torchvision & IN-1k & 83.6 \\
tv\_\allowbreak{}convnext\_\allowbreak{}base & A & \texttt{\scriptsize convnext\_\allowbreak{}base /\allowbreak{} IMAGENET1K\_\allowbreak{}V1} & PyTorch/torchvision & IN-1k & 84.1 \\
tv\_\allowbreak{}convnext\_\allowbreak{}large & A & \texttt{\scriptsize convnext\_\allowbreak{}large /\allowbreak{} IMAGENET1K\_\allowbreak{}V1} & PyTorch/torchvision & IN-1k & 84.4 \\
tv\_\allowbreak{}maxvit\_\allowbreak{}t & A & \texttt{\scriptsize maxvit\_\allowbreak{}t /\allowbreak{} IMAGENET1K\_\allowbreak{}V1} & PyTorch/torchvision & IN-1k & 83.6 \\
timm\_\allowbreak{}resnet50\_\allowbreak{}a1 & A & \texttt{\scriptsize resnet50.\allowbreak{}a1\_\allowbreak{}in1k} & Hugging Face (timm) & IN-1k & 80.4 \\
timm\_\allowbreak{}deit\_\allowbreak{}tiny & A & \texttt{\scriptsize deit\_\allowbreak{}tiny\_\allowbreak{}patch16\_\allowbreak{}224.\allowbreak{}fb\_\allowbreak{}in1k} & Meta (FAIR) & IN-1k & 72.2 \\
timm\_\allowbreak{}deit\_\allowbreak{}small & A & \texttt{\scriptsize deit\_\allowbreak{}small\_\allowbreak{}patch16\_\allowbreak{}224.\allowbreak{}fb\_\allowbreak{}in1k} & Meta (FAIR) & IN-1k & 79.8 \\
timm\_\allowbreak{}deit3\_\allowbreak{}base & A & \texttt{\scriptsize deit3\_\allowbreak{}base\_\allowbreak{}patch16\_\allowbreak{}224.\allowbreak{}fb\_\allowbreak{}in1k} & Meta (FAIR) & IN-1k & 83.8 \\
timm\_\allowbreak{}deit3\_\allowbreak{}large\_\allowbreak{}384\_\allowbreak{}22k & A & \texttt{\scriptsize deit3\_\allowbreak{}large\_\allowbreak{}patch16\_\allowbreak{}384.\allowbreak{}fb\_\allowbreak{}in22k\_\allowbreak{}ft\_\allowbreak{}in1k} & Meta (FAIR) & IN-21k$\to$IN-1k & 87.7 \\
timm\_\allowbreak{}convnext\_\allowbreak{}large\_\allowbreak{}22k & A & \texttt{\scriptsize convnext\_\allowbreak{}large.\allowbreak{}fb\_\allowbreak{}in22k\_\allowbreak{}ft\_\allowbreak{}in1k} & Meta (FAIR) & IN-21k$\to$IN-1k & 86.6 \\
timm\_\allowbreak{}convnextv2\_\allowbreak{}base\_\allowbreak{}22k & A & \texttt{\scriptsize convnextv2\_\allowbreak{}base.\allowbreak{}fcmae\_\allowbreak{}ft\_\allowbreak{}in22k\_\allowbreak{}in1k} & Meta (FAIR) & IN-21k$\to$IN-1k (FCMAE) & 86.7 \\
timm\_\allowbreak{}xcit\_\allowbreak{}medium\_\allowbreak{}p16\_\allowbreak{}384 & A & \texttt{\scriptsize xcit\_\allowbreak{}medium\_\allowbreak{}24\_\allowbreak{}p16\_\allowbreak{}384.\allowbreak{}fb\_\allowbreak{}dist\_\allowbreak{}in1k} & Meta (FAIR) & IN-1k (distilled) & 85.4 \\
timm\_\allowbreak{}cait\_\allowbreak{}s24\_\allowbreak{}384 & A & \texttt{\scriptsize cait\_\allowbreak{}s24\_\allowbreak{}384.\allowbreak{}fb\_\allowbreak{}dist\_\allowbreak{}in1k} & Meta (FAIR) & IN-1k (distilled) & 85.0 \\
timm\_\allowbreak{}levit\_\allowbreak{}384 & A & \texttt{\scriptsize levit\_\allowbreak{}384.\allowbreak{}fb\_\allowbreak{}dist\_\allowbreak{}in1k} & Meta (FAIR) & IN-1k (distilled) & 82.6 \\
timm\_\allowbreak{}resmlp\_\allowbreak{}big\_\allowbreak{}22k & A & \texttt{\scriptsize resmlp\_\allowbreak{}big\_\allowbreak{}24\_\allowbreak{}224.\allowbreak{}fb\_\allowbreak{}in22k\_\allowbreak{}ft\_\allowbreak{}in1k} & Meta (FAIR) & IN-21k$\to$IN-1k & 84.4 \\
timm\_\allowbreak{}mvitv2\_\allowbreak{}large & A & \texttt{\scriptsize mvitv2\_\allowbreak{}large.\allowbreak{}fb\_\allowbreak{}in1k} & Meta (FAIR) & IN-1k & 85.2 \\
timm\_\allowbreak{}hiera\_\allowbreak{}large & A & \texttt{\scriptsize hiera\_\allowbreak{}large\_\allowbreak{}224.\allowbreak{}mae\_\allowbreak{}in1k\_\allowbreak{}ft\_\allowbreak{}in1k} & Meta (FAIR) & IN-1k (MAE) & 86.1 \\
timm\_\allowbreak{}beitv2\_\allowbreak{}large\_\allowbreak{}22k & A & \texttt{\scriptsize beitv2\_\allowbreak{}large\_\allowbreak{}patch16\_\allowbreak{}224.\allowbreak{}in1k\_\allowbreak{}ft\_\allowbreak{}in22k\_\allowbreak{}in1k} & Microsoft & IN-1k$\to$IN-21k$\to$IN-1k & 88.4 \\
timm\_\allowbreak{}swinv2\_\allowbreak{}large\_\allowbreak{}384\_\allowbreak{}22k & A & \texttt{\scriptsize swinv2\_\allowbreak{}large\_\allowbreak{}window12to24\_\allowbreak{}192to384.\allowbreak{}ms\_\allowbreak{}in22k\_\allowbreak{}ft\_\allowbreak{}in1k} & Microsoft & IN-21k$\to$IN-1k & 87.5 \\
timm\_\allowbreak{}tiny\_\allowbreak{}vit\_\allowbreak{}21m\_\allowbreak{}512 & A & \texttt{\scriptsize tiny\_\allowbreak{}vit\_\allowbreak{}21m\_\allowbreak{}512.\allowbreak{}dist\_\allowbreak{}in22k\_\allowbreak{}ft\_\allowbreak{}in1k} & Microsoft & IN-21k$\to$IN-1k (distilled) & 86.4 \\
timm\_\allowbreak{}fastvit\_\allowbreak{}ma36 & A & \texttt{\scriptsize fastvit\_\allowbreak{}ma36.\allowbreak{}apple\_\allowbreak{}dist\_\allowbreak{}in1k} & Apple & IN-1k (distilled) & 84.6 \\
timm\_\allowbreak{}mobileone\_\allowbreak{}s4 & A & \texttt{\scriptsize mobileone\_\allowbreak{}s4.\allowbreak{}apple\_\allowbreak{}in1k} & Apple & IN-1k & 79.4 \\
timm\_\allowbreak{}gcvit\_\allowbreak{}base & A & \texttt{\scriptsize gcvit\_\allowbreak{}base.\allowbreak{}in1k} & NVIDIA & IN-1k & 84.5 \\
timm\_\allowbreak{}rexnet\_\allowbreak{}200 & A & \texttt{\scriptsize rexnet\_\allowbreak{}200.\allowbreak{}nav\_\allowbreak{}in1k} & NAVER & IN-1k & 81.6 \\
timm\_\allowbreak{}pit\_\allowbreak{}b & A & \texttt{\scriptsize pit\_\allowbreak{}b\_\allowbreak{}224.\allowbreak{}in1k} & NAVER & IN-1k & 82.4 \\
timm\_\allowbreak{}crossvit\_\allowbreak{}18\_\allowbreak{}dagger\_\allowbreak{}408 & A & \texttt{\scriptsize crossvit\_\allowbreak{}18\_\allowbreak{}dagger\_\allowbreak{}408.\allowbreak{}in1k} & IBM & IN-1k & 84.2 \\
timm\_\allowbreak{}convmixer\_\allowbreak{}1536\_\allowbreak{}20 & A & \texttt{\scriptsize convmixer\_\allowbreak{}1536\_\allowbreak{}20.\allowbreak{}in1k} & Carnegie Mellon University & IN-1k & 81.4 \\
timm\_\allowbreak{}efficientformerv2\_\allowbreak{}l & A & \texttt{\scriptsize efficientformerv2\_\allowbreak{}l.\allowbreak{}snap\_\allowbreak{}dist\_\allowbreak{}in1k} & Snap & IN-1k (distilled) & 83.4 \\
timm\_\allowbreak{}caformer\_\allowbreak{}b36\_\allowbreak{}22k & A & \texttt{\scriptsize caformer\_\allowbreak{}b36.\allowbreak{}sail\_\allowbreak{}in22k\_\allowbreak{}ft\_\allowbreak{}in1k} & Sea AI Lab & IN-21k$\to$IN-1k & 87.4 \\
timm\_\allowbreak{}volo\_\allowbreak{}d5\_\allowbreak{}512 & A & \texttt{\scriptsize volo\_\allowbreak{}d5\_\allowbreak{}512.\allowbreak{}sail\_\allowbreak{}in1k} & Sea AI Lab & IN-1k & 87.1 \\
timm\_\allowbreak{}dla102x2 & A & \texttt{\scriptsize dla102x2.\allowbreak{}in1k} & UC Berkeley & IN-1k & 79.4 \\
timm\_\allowbreak{}selecsls60b & A & \texttt{\scriptsize selecsls60b.\allowbreak{}in1k} & Max Planck Institute for Informatics & IN-1k & 78.4 \\
timm\_\allowbreak{}edgenext\_\allowbreak{}base & A & \texttt{\scriptsize edgenext\_\allowbreak{}base.\allowbreak{}in21k\_\allowbreak{}ft\_\allowbreak{}in1k} & MBZUAI & IN-21k$\to$IN-1k & 83.4 \\
clip\_\allowbreak{}openai\_\allowbreak{}rn50 & B & \texttt{\scriptsize RN50-quickgelu /\allowbreak{} openai} & OpenAI & WIT-400M & 59.7 \\
clip\_\allowbreak{}openai\_\allowbreak{}rn101 & B & \texttt{\scriptsize RN101-quickgelu /\allowbreak{} openai} & OpenAI & WIT-400M & 62.1 \\
clip\_\allowbreak{}openai\_\allowbreak{}rn50x16 & B & \texttt{\scriptsize RN50x16-quickgelu /\allowbreak{} openai} & OpenAI & WIT-400M & 70.7 \\
clip\_\allowbreak{}openai\_\allowbreak{}rn50x64 & B & \texttt{\scriptsize RN50x64-quickgelu /\allowbreak{} openai} & OpenAI & WIT-400M & 73.9 \\
clip\_\allowbreak{}openai\_\allowbreak{}vitb32 & B & \texttt{\scriptsize ViT-B-32-quickgelu /\allowbreak{} openai} & OpenAI & WIT-400M & 63.2 \\
clip\_\allowbreak{}openai\_\allowbreak{}vitb16 & B & \texttt{\scriptsize ViT-B-16-quickgelu /\allowbreak{} openai} & OpenAI & WIT-400M & 68.3 \\
clip\_\allowbreak{}openai\_\allowbreak{}vitl14 & B & \texttt{\scriptsize ViT-L-14-quickgelu /\allowbreak{} openai} & OpenAI & WIT-400M & 75.5 \\
clip\_\allowbreak{}openai\_\allowbreak{}vitl14\_\allowbreak{}336 & B & \texttt{\scriptsize ViT-L-14-336-quickgelu /\allowbreak{} openai} & OpenAI & WIT-400M & 76.6 \\
clip\_\allowbreak{}laion2b\_\allowbreak{}vitb32 & B & \texttt{\scriptsize ViT-B-32 /\allowbreak{} laion2b\_\allowbreak{}s34b\_\allowbreak{}b79k} & LAION & LAION-2B & 66.4 \\
clip\_\allowbreak{}laion2b\_\allowbreak{}vitl14 & B & \texttt{\scriptsize ViT-L-14 /\allowbreak{} laion2b\_\allowbreak{}s32b\_\allowbreak{}b82k} & LAION & LAION-2B & 75.3 \\
clip\_\allowbreak{}laion2b\_\allowbreak{}vith14 & B & \texttt{\scriptsize ViT-H-14 /\allowbreak{} laion2b\_\allowbreak{}s32b\_\allowbreak{}b79k} & LAION & LAION-2B & 77.9 \\
clip\_\allowbreak{}laion2b\_\allowbreak{}vitg14 & B & \texttt{\scriptsize ViT-bigG-14 /\allowbreak{} laion2b\_\allowbreak{}s39b\_\allowbreak{}b160k} & LAION & LAION-2B & 80.1 \\
clip\_\allowbreak{}laion2b\_\allowbreak{}convnext\_\allowbreak{}xxl & B & \texttt{\scriptsize convnext\_\allowbreak{}xxlarge /\allowbreak{} laion2b\_\allowbreak{}s34b\_\allowbreak{}b82k\_\allowbreak{}augreg\_\allowbreak{}soup} & LAION & LAION-2B & 79.4 \\
clip\_\allowbreak{}datacomp\_\allowbreak{}vitl14 & B & \texttt{\scriptsize ViT-L-14 /\allowbreak{} datacomp\_\allowbreak{}xl\_\allowbreak{}s13b\_\allowbreak{}b90k} & DataComp (LAION/UW consortium) & DataComp-1B & 79.1 \\
clip\_\allowbreak{}dfn2b\_\allowbreak{}vitl14 & B & \texttt{\scriptsize ViT-L-14-quickgelu /\allowbreak{} dfn2b} & Apple & DFN-2B & 81.4 \\
clip\_\allowbreak{}dfn5b\_\allowbreak{}vith14 & B & \texttt{\scriptsize ViT-H-14-quickgelu /\allowbreak{} dfn5b} & Apple & DFN-5B & 83.4 \\
clip\_\allowbreak{}metaclip\_\allowbreak{}vitb16 & B & \texttt{\scriptsize ViT-B-16-quickgelu /\allowbreak{} metaclip\_\allowbreak{}fullcc} & Meta (FAIR) & MetaCLIP-2.5B & 72.1 \\
clip\_\allowbreak{}metaclip\_\allowbreak{}vitl14 & B & \texttt{\scriptsize ViT-L-14-quickgelu /\allowbreak{} metaclip\_\allowbreak{}fullcc} & Meta (FAIR) & MetaCLIP-2.5B & 79.1 \\
clip\_\allowbreak{}metaclip\_\allowbreak{}vith14 & B & \texttt{\scriptsize ViT-H-14-quickgelu /\allowbreak{} metaclip\_\allowbreak{}fullcc} & Meta (FAIR) & MetaCLIP-2.5B & 80.5 \\
clip\_\allowbreak{}pe\_\allowbreak{}core\_\allowbreak{}b16 & B & \texttt{\scriptsize PE-Core-B-16 /\allowbreak{} meta} & Meta (FAIR) & MetaCLIP-5.4B (PE) & 78.4 \\
clip\_\allowbreak{}pe\_\allowbreak{}core\_\allowbreak{}l14\_\allowbreak{}336 & B & \texttt{\scriptsize PE-Core-L-14-336 /\allowbreak{} meta} & Meta (FAIR) & MetaCLIP-5.4B (PE) & 83.4 \\
clip\_\allowbreak{}pe\_\allowbreak{}core\_\allowbreak{}bigg14\_\allowbreak{}448 & B & \texttt{\scriptsize PE-Core-bigG-14-448 /\allowbreak{} meta} & Meta (FAIR) & MetaCLIP-5.4B (PE) & 85.3 \\
tv\_\allowbreak{}regnet\_\allowbreak{}y\_\allowbreak{}16gf\_\allowbreak{}swag & C & \texttt{\scriptsize regnet\_\allowbreak{}y\_\allowbreak{}16gf /\allowbreak{} IMAGENET1K\_\allowbreak{}SWAG\_\allowbreak{}E2E\_\allowbreak{}V1} & Meta (FAIR) & IG-3.6B$\to$IN-1k & 86.0 \\
tv\_\allowbreak{}vit\_\allowbreak{}b\_\allowbreak{}16\_\allowbreak{}swag & C & \texttt{\scriptsize vit\_\allowbreak{}b\_\allowbreak{}16 /\allowbreak{} IMAGENET1K\_\allowbreak{}SWAG\_\allowbreak{}E2E\_\allowbreak{}V1} & Meta (FAIR) & IG-3.6B$\to$IN-1k & 85.3 \\
tv\_\allowbreak{}vit\_\allowbreak{}l\_\allowbreak{}16\_\allowbreak{}swag & C & \texttt{\scriptsize vit\_\allowbreak{}l\_\allowbreak{}16 /\allowbreak{} IMAGENET1K\_\allowbreak{}SWAG\_\allowbreak{}E2E\_\allowbreak{}V1} & Meta (FAIR) & IG-3.6B$\to$IN-1k & 88.1 \\
tv\_\allowbreak{}vit\_\allowbreak{}h\_\allowbreak{}14\_\allowbreak{}swag & C & \texttt{\scriptsize vit\_\allowbreak{}h\_\allowbreak{}14 /\allowbreak{} IMAGENET1K\_\allowbreak{}SWAG\_\allowbreak{}E2E\_\allowbreak{}V1} & Meta (FAIR) & IG-3.6B$\to$IN-1k & 88.6 \\
timm\_\allowbreak{}resnext101\_\allowbreak{}32x16d\_\allowbreak{}wsl & C & \texttt{\scriptsize resnext101\_\allowbreak{}32x16d.\allowbreak{}fb\_\allowbreak{}wsl\_\allowbreak{}ig1b\_\allowbreak{}ft\_\allowbreak{}in1k} & Meta (FAIR) & IG-940M$\to$IN-1k & 84.2 \\
timm\_\allowbreak{}resnext101\_\allowbreak{}32x8d\_\allowbreak{}swsl & C & \texttt{\scriptsize resnext101\_\allowbreak{}32x8d.\allowbreak{}fb\_\allowbreak{}swsl\_\allowbreak{}ig1b\_\allowbreak{}ft\_\allowbreak{}in1k} & Meta (FAIR) & IG-940M semi-weak$\to$IN-1k & 84.3 \\
timm\_\allowbreak{}regnety\_\allowbreak{}640\_\allowbreak{}seer & C & \texttt{\scriptsize regnety\_\allowbreak{}640.\allowbreak{}seer\_\allowbreak{}ft\_\allowbreak{}in1k} & Meta (FAIR) & IG-1B SSL$\to$IN-1k & 83.9 \\
hub\_\allowbreak{}dinov2\_\allowbreak{}vits14\_\allowbreak{}lc & C & \texttt{\scriptsize facebookresearch/\allowbreak{}dinov2 /\allowbreak{} dinov2\_\allowbreak{}vits14\_\allowbreak{}lc} & Meta (FAIR) & LVD-142M SSL + IN-1k linear & 81.4 \\
hub\_\allowbreak{}dinov2\_\allowbreak{}vitb14\_\allowbreak{}lc & C & \texttt{\scriptsize facebookresearch/\allowbreak{}dinov2 /\allowbreak{} dinov2\_\allowbreak{}vitb14\_\allowbreak{}lc} & Meta (FAIR) & LVD-142M SSL + IN-1k linear & 84.5 \\
hub\_\allowbreak{}dinov2\_\allowbreak{}vitl14\_\allowbreak{}lc & C & \texttt{\scriptsize facebookresearch/\allowbreak{}dinov2 /\allowbreak{} dinov2\_\allowbreak{}vitl14\_\allowbreak{}lc} & Meta (FAIR) & LVD-142M SSL + IN-1k linear & 86.4 \\
hub\_\allowbreak{}dinov2\_\allowbreak{}vitg14\_\allowbreak{}lc & C & \texttt{\scriptsize facebookresearch/\allowbreak{}dinov2 /\allowbreak{} dinov2\_\allowbreak{}vitg14\_\allowbreak{}lc} & Meta (FAIR) & LVD-142M SSL + IN-1k linear & 86.5 \\
timm\_\allowbreak{}vitl14\_\allowbreak{}clip\_\allowbreak{}openai\_\allowbreak{}ft & C & \texttt{\scriptsize vit\_\allowbreak{}large\_\allowbreak{}patch14\_\allowbreak{}clip\_\allowbreak{}336.\allowbreak{}openai\_\allowbreak{}ft\_\allowbreak{}in12k\_\allowbreak{}in1k} & OpenAI pretrain; fine-tune Hugging Face (timm) & WIT-400M$\to$IN-12k$\to$IN-1k & 88.3 \\
timm\_\allowbreak{}vith14\_\allowbreak{}clip\_\allowbreak{}laion\_\allowbreak{}ft & C & \texttt{\scriptsize vit\_\allowbreak{}huge\_\allowbreak{}patch14\_\allowbreak{}clip\_\allowbreak{}336.\allowbreak{}laion2b\_\allowbreak{}ft\_\allowbreak{}in12k\_\allowbreak{}in1k} & LAION pretrain; fine-tune Hugging Face (timm) & LAION-2B$\to$IN-12k$\to$IN-1k & 88.6 \\
timm\_\allowbreak{}convnext\_\allowbreak{}xxl\_\allowbreak{}clip\_\allowbreak{}laion\_\allowbreak{}ft & C & \texttt{\scriptsize convnext\_\allowbreak{}xxlarge.\allowbreak{}clip\_\allowbreak{}laion2b\_\allowbreak{}soup\_\allowbreak{}ft\_\allowbreak{}in1k} & LAION pretrain; fine-tune Hugging Face (timm) & LAION-2B$\to$IN-1k & 88.6 \\
\bottomrule
\end{longtable}
\end{small}
\end{document}